\documentclass[runningheads, envcountsame, a4paper]{llncs} %

\usepackage[utf8]{inputenc}
\usepackage[T1]{fontenc}
\usepackage{microtype}      %
\usepackage{csquotes}
\usepackage{siunitx}
\usepackage[hyphens]{url}            %
\usepackage{booktabs}       %
\usepackage{amsfonts}       %
\usepackage{nicefrac}       %
\usepackage{amsmath}
\usepackage{amssymb}
\usepackage{bm}
\usepackage[misc]{ifsym}
\usepackage{chngcntr}

\usepackage[pdftex]{graphicx}
\usepackage[table]{xcolor}
\usepackage{pifont}  %
\usepackage{hyperref}

\def\bbbone{{\mathchoice {\rm 1\mskip-4mu l} {\rm 1\mskip-4mu l}
		{\rm 1\mskip-4.5mu l} {\rm 1\mskip-5mu l}}}

\def\ourmodel{{ALMGIG}}

\newcommand{\cmark}{\ding{51}}%
\newcommand{\xmark}{\ding{55}}%

\title{Adversarial Learned Molecular Graph Inference and Generation}
\begin{document} %
\titlerunning{Adversarial Molecular Graph Inference and Generation}
\author{Sebastian P{\"{o}}lsterl$^\text{(\Letter)}$ \and
   Christian Wachinger}
\authorrunning{S.~P{\"{o}}lsterl and C.~Wachinger} %
\toctitle{Adversarial Learned Molecular Graph Inference and Generation}%
\tocauthor{Sebastian~P{\"{o}}lsterl and Christian~Wachinger}%
\institute{Artificial Intelligence in Medical Imaging (AI-Med),\\
   Department of Child and Adolescent Psychiatry,\\
   Ludwig-Maximilians-Universit{\"{a}}t, Munich, Germany\\
   \email{sebastian.poelsterl@med.uni-muenchen.de}} %
\maketitle              %

\begin{abstract}
	Recent methods for generating novel molecules use
	graph representations of molecules and employ various forms of graph convolutional
	neural networks for inference. However, training requires solving an
	expensive graph isomorphism problem, which previous approaches
	do not address or solve only approximately.
	In this work, we propose ALMGIG, a likelihood-free adversarial learning framework
	for inference and \textit{de novo} molecule generation that avoids explicitly computing a reconstruction loss.
	Our approach extends generative adversarial networks
	by including an adversarial cycle-consistency loss to
	implicitly enforce the reconstruction property.
	To capture properties unique to molecules, such as valence,
	we extend the Graph Isomorphism Network to multi-graphs.
	To quantify the performance of models, we propose to compute the
	distance between distributions of physicochemical properties with the
	1-Wasserstein distance.
	We demonstrate that ALMGIG
	more accurately learns
	the distribution over the space of molecules than all baselines.
	Moreover, it can be utilized for drug discovery by efficiently searching the space of molecules
	using molecules' continuous latent representation.
	Our code is available at \url{https://github.com/ai-med/almgig}
\end{abstract}

\section{Introduction}

Deep generative models have been proven successful in generating high-quality
samples in the domain of images, audio, and text, but it was only recently
when models have been developed for \textit{de novo} chemical design
\cite{Gomez-Bombarelli2018,Kusner2017}.
The goal of \textit{de novo} chemical design is to map desirable properties of molecules, such
as a drug being active against a certain biological target, to the space of
molecules. This process -- called inverse Quantitative Structure-Activity Relationship (QSAR) --
is extremely challenging due to the vast size of the chemical space, which is
estimated to contain in the order of $10^{33}$ drug-like molecules~\cite{Polishchuk2013}.
Searching this space efficiently is often hindered by the discrete nature of molecules,
which prevents the use of gradient-based optimization.
Thus, obtaining a continuous and differentiable representation of molecules
is a desirable goal that could ease drug discovery.
For \textit{de novo} generation of molecules,
it is important to produce chemically valid molecules
that comply with the valence of atoms, i.e., how many electron pairs
an atom of a particular type can share.
For instance, carbon has a valence of four and can form at most four single bonds.
Therefore, any mapping from the continuous latent space of a model to the space
of molecules should result in a chemically valid molecule.

The current state-of-the-art deep learning models are
adversarial or variational autoencoders (AAE, VAE)
that represent molecules as graphs and rely on graph convolutional neural networks (GCNs)
\cite{DeCao2018,Jin2018,Li2018a,Li2018,Liu2018,Ma2018,Samanta2018,Simonovsky2018,You2018}.
The main obstacle is in defining a suitable reconstruction loss, which
is challenging when inputs and outputs are graphs.
Because there is no canonical form of a graph's adjacency matrix,
two graphs can be identical despite having different
adjacency matrices.
Before the reconstruction loss can be computed, correspondences
between nodes of the target and reconstructed graph
need to be established,
which requires solving a
computationally expensive graph isomorphism problem.
Existing graph-based VAEs have addressed this problem by
either traversing nodes in a fixed order \cite{Jin2018,Samanta2018,Liu2018}
or employing graph matching algorithms \cite{Simonovsky2018} to approximate the
reconstruction loss.

\begin{figure*}[tb]
	\centering
	\includegraphics[width=\linewidth]{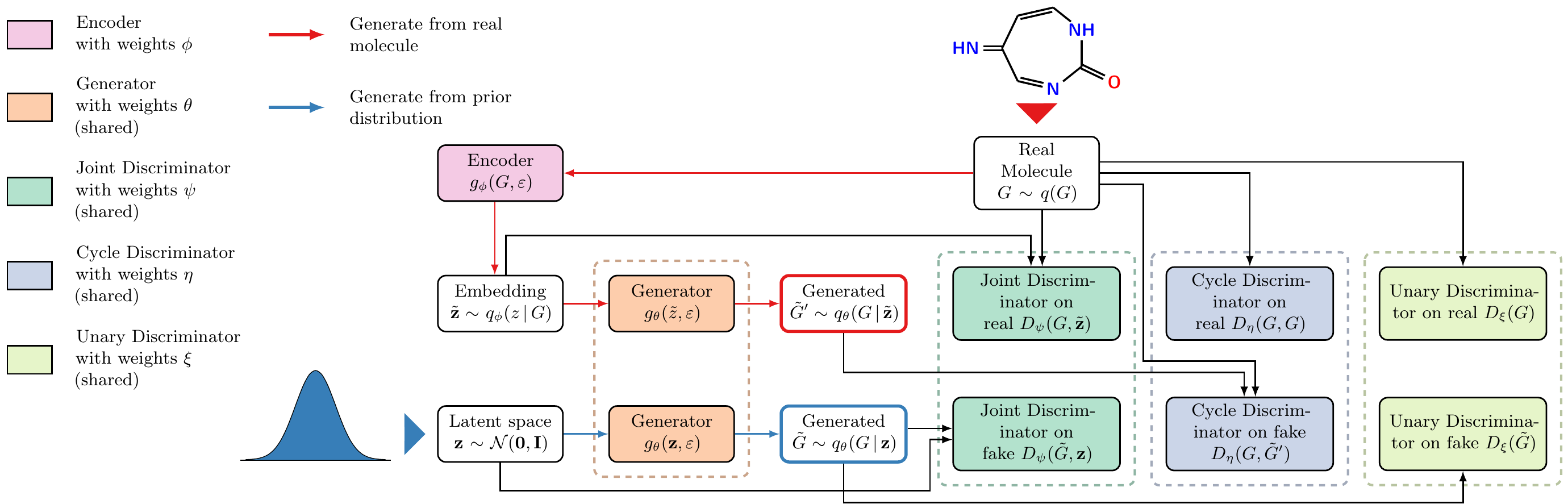}
	\caption{\label{fig:model-overview}%
		Overview of the proposed \ourmodel\ model. Boxes with identical background color
		represent neural networks that share their weights.
		The joint discriminator plays a similar role as the discriminator
		in standard GANs. The cycle discriminator enforces the reconstruction
		property without explicitly computing a reconstruction loss.
		Molecules can be generated by transforming a sample from a simple prior
		distribution (blue path), or by embedding a real molecule into the latent space and
		reconstructing it (red path).}
\end{figure*}

We propose \ourmodel, a likelihood-free Generative Adversarial Network for inference and generation
of molecular graphs (see fig. \ref{fig:model-overview}).
This is the first time that an inference and generative model of molecular graphs
can be trained without computing a reconstruction loss.
Our model consists of an encoder (inference model) and a decoder (generator)
that are trained by implicitly imposing the reconstructing property
via cycle-consistency, thus, avoiding the need to solve a
computationally prohibitive graph isomorphism problem.
To learn from graph-structered data, we base
our encoder on the recently proposed Graph Isomorphism Network
\cite{Xu2019}, which we extend to multi-graphs,
and employ the Gumbel-softmax trick \cite{Jang2017,Maddison2017}
to generate discrete molecular graphs.
Finally, we explicitly incorporate domain knowledge
such that generated graphs represent
valid chemical structures.
We will show that this enables us to perform efficient nearest
neighbor search in the space of molecules.

In addition, %
we performed an extensive suite of benchmarks to accurately
determine the strengths and weaknesses of models.
We argue that
summary statistics such as the percentage of
valid, unique, and novel molecules used in previous studies,
are poor proxies to determine whether
generated molecules are chemically meaningful.
We instead compare the distributions of 10
chemical properties and demonstrate that our proposed
method is able to more accurately learn a distribution over
the space of molecules than previous approaches.
\section{Related Work}

Graphs, where nodes represent atoms, and edges chemical bonds,
are a natural representation of molecules, which has been
explored in
\cite{DeCao2018,Jin2018,Li2018a,Li2018,Liu2018,Ma2018,Samanta2018,Simonovsky2018,You2018}.
Most methods rely on graph convolutional neural networks (GCNs)
for inference, which can efficiently learn from the non-Euclidean structure of graphs
\cite{Bradshaw2019,DeCao2018,Li2018a,Liu2018,Samanta2018,Simonovsky2018,You2018}.
Molecular graphs can be generated sequentially, adding single atoms
or small fragments using an RNN-based architecture
\cite{Bradshaw2019,Jin2018,Li2018a,Li2018,Liu2018,Podda2020,Samanta2018,You2018},
or in a single step \cite{DeCao2018,Ma2018,Simonovsky2018}.
Sequential generation has the advantage that partially generated graphs
can be checked, e.g., for valence violations \cite{Liu2018,Samanta2018}.
Molecules can also be represented as strings using
SMILES encoding, for which previous work relied on recurrent neural networks
for inference and generation
\cite{Blaschke2018,Dai2018,Gomez-Bombarelli2018,Guimaraes2017,Kusner2017,Lim2018,Olivecrona2017,Podda2020,Popova2018,Putin2018,Segler2018}.
However, producing valid SMILES strings is challenging, because
models need to learn the underlying grammar of SMILES.
Therefore, a considerable portion of generated SMILES tend
to be invalid (15-80\%) \cite{Blaschke2018,Guimaraes2017,Lim2018,Putin2018,Segler2018}
-- unless constraints are built into the model
\cite{Dai2018,Kusner2017,Li2018a,Popova2018}.
The biggest downside of the SMILES representation is that it
does not capture molecular similarity: substituting a single
character can alter the underlying molecule structure significantly
or invalidate it. Therefore, partially generated SMILES strings
cannot be validated and transitions in the latent space of such models
may lead to abrupt changes in molecule structure \cite{Jin2018}. %

With respect to the generative model,
most previous work use either VAEs or
adversarial learning.
VAEs and AAEs take a molecule representation as
input and project it via an inference model (encoder) to a latent space, which is subsequently
transformed by a decoder to produce a molecule representation
\cite{Blaschke2018,Dai2018,Gomez-Bombarelli2018,Jin2018,Kadurin2017,Kusner2017,Lim2018,Liu2018,Samanta2018,Simonovsky2018}.
New molecules can be generated by drawing points from
a simple prior distribution over the latent space (usually Gaussian)
and feeding it to the decoder.
AAEs \cite{Blaschke2018,Kadurin2017} perform variational
inference by adversarial learning, using a discriminator
as in GANs, which
allows using a more complex prior distribution
over the latent space.
Both VAEs and AAEs are trained to minimize a reconstruction
loss, which is expensive to compute.
Standard GANs for molecule generation lack an encoder and a reconstruction loss,
and are trained via a two-player game
between a generator and discriminator \cite{DeCao2018,Guimaraes2017,Putin2018}.
The generator transforms a simple input distribution
into a distribution in the space of molecules,
such that the discriminator is unable to distinguish
between real molecules in the training data and generated molecules.
Without an inference model, GANs cannot be used for
neighborhood search in the latent space.
\section{Methods}

We represent molecules as graphs and propose
our Adversarial Learned Molecular Graph Inference and Generation
(\ourmodel) framework
to learn a distribution
in the space of molecules, generate novel molecules, and
efficiently search the neighborhood of existing molecules.
\ourmodel\ is a bidirectional GAN that has a GCN-based encoder that projects a graph
into the latent space and a decoder that outputs
a one-hot representation of atoms and an adjacency matrix
defining atomic bonds.
Hence, graph generation is performed in a single
step, which allows considering global properties of a molecule.
An undirected multi-graph $G = (\mathcal{V}, \mathcal{R}, \mathcal{E})$
is defined by its vertices
$\mathcal{V} = \{v_1, \ldots, v_n \}$,
relation types $\mathcal{R} = \{r_1, \ldots, r_m \}$,
and typed edges (relations)
$\mathcal{E} = \{ (v_i, r_k, v_j)\,|\,v_i,v_j \in \mathcal{V}, r_k \in \mathcal{R} \}$.
Here, we only allow vertices to be connected by at most one type of edge.
We represent a multi-graph $G$ by its adjacency tensor $\mathbf{A} \in \{0, 1\}^{n \times n \times m}$,
where $A_{ijk}$ is one if $(v_i, r_k, v_j) \in \mathcal{E}$ and zero otherwise,
and the node feature matrix
$\mathbf{X} = \left( \mathbf{x}_{v_1}, \ldots, \mathbf{x}_{v_n} \right)^\top
\in \mathbb{R}^{n \times d}$, where $d$ is the number of features describing
each node.
Here, vertices are atoms, $\mathcal{R}$ is the set of bonds considered
(single, double, triple), and node feature vectors $\mathbf{x}_{v_i}$ are
one-hot encoded atom types (carbon, oxygen, …), with $d$ representing
the total number of atom types.
We do not model hydrogen atoms, but implicitly add hydrogens
to match an atom's valence.

\subsection{Adversarially Learned Inference}

We first describe Adversarially Learned Inference with Conditional Entropy (ALICE) \cite{Li2017}, which is at the core of
\ourmodel\ and has not been used for \textit{de novo} chemical design before.
It allows training our encoder-decoder model fully
adversarially and implicitly enforces the reconstruction property
without the need to compute a reconstruction loss
(see overview in fig.~\ref{fig:model-overview}).
It has two generative models: (i) the \emph{encoder} maps graphs
into the latent space, and (ii) the \emph{decoder} inverts this transformation
by mapping latent codes to graphs.
Training is performed by matching joint distributions
over graphs $G \in \mathcal{G}$ and latent
variables $\mathbf{z} \in \mathcal{Z}$. Let $q(G)$ be
the marginal data distribution, from which we have samples,
$\tilde{\mathbf{z}}$
a latent representation produced by the encoder, and
$\tilde{G}$ a generated graph.
The encoder generative model over the latent variables is $q_\phi(z\,|\,G)$
with parameters $\phi$, and the decoder generative model over graphs
is $q_\theta(G\,|\,z)$, parametrized by $\theta$.
Putting everything together, we obtain the \emph{encoder joint distribution}
$q_\phi(G, z) = q(G) q_\phi(z\,|\,G)$, and
the \emph{decoder joint distribution}
$p_\theta(G, z) = p_z(z) q_\theta(G\,|\,z)$.
The objective of Adversarially Learned Inference is to match the two joint
distributions~\cite{Dumoulin2017}.
A discriminator network $D_\psi$ with
parameters $\psi$ is trained to distinguish samples
$(G, \tilde{\mathbf{z}}) \sim q_\phi(G, z)$ from $(\tilde{G}, \mathbf{z}) \sim p_\theta(G, z)$.
Drawing samples $\tilde{\mathbf{z}}$ and $\tilde{G}$
is made possible by specifying the encoder $q_\phi$ and decoder $q_\theta$
as neural networks using the change of variable technique:
$\tilde{\mathbf{z}} = g_\phi(G, \varepsilon)$, $\tilde{G} = g_\theta(\mathbf{z}, \varepsilon)$,
where $\varepsilon$ is some random source of noise.
The objective then becomes the following min-max game:
\begin{equation}\label{eq:ali}
\begin{split}
\min_{\theta,\phi} \max_\psi \quad&%
\mathbb{E}_{G \sim q(G),~ \tilde{\mathbf{z}} \sim q_\phi(z\,|\,G)} [ \log \sigma( D_\psi(G, \tilde{\mathbf{z}})) ] \\
&+ \mathbb{E}_{\tilde{G} \sim q_\theta(G\,|\,z),~ \mathbf{z} \sim p(z)} [ \log(1 - \sigma(D_\psi(\tilde{G}, \mathbf{z})) ) ] .
\end{split}
\end{equation}
where $\sigma(\cdot)$ denotes the sigmoid function.
At the optimum of \eqref{eq:ali},
the marginal distributions
$q_\theta(G\,|\,z)$ and $q_\phi(z\,|\,G)$ will match, but their relationship
can still be undesirable.
For instance, the
decoder $g_\theta(\mathbf{z}, \varepsilon)$ could map a given $z$ to $G_1$ half of
the time and to a distinct $G_2$ the other half of the time
-- the same applies to the encoder~\cite{Li2017}.
ALICE \cite{Li2017}
solves this issue by including a cycle-consistency constraint via
an additional adversarial loss.
This encourages encoder and decoder to mimic the reconstruction property
without explicitly solving a computationally demanding graph isomorphism problem.
To this end, a second discriminator with parameters $\eta$ is
trained to distinguish a real from a reconstructed graph:
\begin{equation}\label{eq:alice-cycle}
\begin{split}
\min_{\theta,\phi} \max_\eta \quad&%
\mathbb{E}_{G \sim q(G)} [ \log \sigma(D_\eta(G, G)) ] \\
&+ \mathbb{E}_{\tilde{G}^\prime \sim q_\theta(G\,|\,\tilde{z}),~ \tilde{\mathbf{z}} \sim q_\phi(z\,|\,G)}
[ \log(1 - \sigma(D_\eta(G, \tilde{G}^\prime))) ] .
\end{split}
\end{equation}

While encoder and decoder in ALICE have all the desired properties at the optimum,
we rarely find the global optimum, because both are deep neural networks.
Therefore, we extend ALICE by including a unary discriminator $D_\xi$ only
on graphs, as in standard GANs.
This additional objective facilitates generator training when the
joint distribution is difficult to learn:
\begin{equation}\label{eq:alice-unary}
	\begin{split}
	\min_{\theta,\phi} \max_\xi \quad&%
	\mathbb{E}_{G \sim q(G)} [ \log \sigma( D_\xi(G)) ] \\
	&+ \mathbb{E}_{\tilde{G} \sim q_\theta(G\,|\,z),~ \mathbf{z} \sim p(z)} [ \log(1 - \sigma(D_\xi(\tilde{G})) ) ] .
	\end{split}
	\end{equation}
We will demonstrate in our ablation study that this is indeed essential.

During training, we concurrently optimize
\eqref{eq:ali}, \eqref{eq:alice-cycle}, and \eqref{eq:alice-unary}
by first updating $\theta$ and $\phi$, while keeping $\psi$, $\eta$ and $\xi$ fixed,
and then the other way around.
We use a higher learning rate when updating encoder and decoder weights $\theta$, $\phi$
as proposed in~\cite{Heusel2017}.
Finally, we employ the 1-Lipschitz constraint in \cite{Gulrajani2017}
such that discriminators $D_\psi$, $D_\eta$, and $D_\xi$ are approximately
1-Lipschtiz continuous.
Next, we will describe the architecture of the encoder, decoder, and discriminators.

\subsection{Generator}

Adjacency matrix (bonds) and node types (atoms) of a molecular graph are discrete
structures.
Therefore, the generator must define an implicit discrete distribution
over edges and node types, which differs from
traditional GANs that can only model continuous distributions.
We overcome this issue by using the Gumbel-softmax trick
similar to~\cite{DeCao2018}.
The generator network $g_\theta(\mathbf{z}, \varepsilon)$ takes a point $\mathbf{z}$ from latent space and noise $\varepsilon$,
and outputs a discrete-valued and symmetric graph adjacency tensor~$\mathbf{A}$ and
a discrete-valued node feature matrix~$\mathbf{X}$.
We use an MLP with three hidden layers with 128, 256, 512 units
and $\tanh$ activation, respectively.
To facilitate better gradient flow, we employ skip-connections from
$\mathbf{z}$ to layers of the MLP.
First, the latent vector $\mathbf{z}$ is split into three equally-sized parts
$\mathbf{z}_1, \mathbf{z}_2, \mathbf{z}_3$.
The input to the first hidden layer is the concatenation
$[\mathbf{z}_1, \varepsilon]$, its output is concatenated
with $\mathbf{z}_2$ and fed to the second layer,
and similarly for the third layer with $\mathbf{z}_3$.

We extend $\mathbf{A}$ and $\mathbf{X}$ to explicitly
model the absence of edges and nodes by introducing
a separate ghost-edge type and ghost-node type.
This will enable us to encourage the generator to produce
chemically valid molecular graphs as described below.
Thus, we define $\tilde{\mathbf{A}} \in \{0, 1\}^{n \times n \times (m+1)}$
and $\tilde{\mathbf{X}} \in \{0, 1\}^{n \times (d +1)}$.
Each vector $\tilde{\mathbf{A}}_{ij\bullet}$, representing generated
edges between nodes $i$ and $j$, needs to be a member of the simplex
$\Delta^{m} = \{ (y_0, y_1,\ldots,y_m)\,|\, y_k \in \{0,1\}, \sum_{k=0}^m y_k = 1 \}$,
because only none or a single edge between $i$ and $j$ is allowed.
Here, we use the zero element to represent the absence of an edge.
Similarly, each generated node feature vector $\tilde{\mathbf{x}}_{v_i}$
needs to be a member of the simplex $\Delta^d$, where the zero
element represents ghost nodes.

\paragraph{Gumbel-softmax Trick.}
The generator is a neural network with two outputs,
$\mathrm{MLP}_A(\mathbf{z}) \in \mathbb{R}^{n \times n \times (m+1)}$
and
$\mathrm{MLP}_X(\mathbf{z}) \in \mathbb{R}^{n \times (d+1)}$,
which are created by linearly projecting hidden units into
a $n^2 (m+1)$ and $n(d+1)$ dimensional space, respectively.
Next, continuous outputs need to be transformed
into discrete values according to the rules above
to obtain tensors $\tilde{\mathbf{A}}$ and $\tilde{\mathbf{X}}$
representing a generated graph.
Since the argmax operation is non-differentiable,
we employ the Gumbel-softmax trick~\cite{Jang2017,Maddison2017},
which uses reparameterization to obtain a continuous relaxation of discrete states.
Thus, we obtain an approximately discrete
adjacency tensor $\tilde{\mathbf{A}}$ from $\mathrm{MLP}_A(\mathbf{z})$,
and feature matrix $\tilde{\mathbf{X}}$ from $\mathrm{MLP}_X(\mathbf{z})$.

\paragraph{Node Connectivity and Valence Constraints.}

While this allows us to generate graphs with varying number of nodes,
the generator could in principle generate graphs consisting
of two or more separate connected components.
In addition, generating molecules where atoms have the correct
number of shared electron pairs (valence) is an important
aspect the generator needs to consider, otherwise generated
graphs would represent invalid molecules.
Finally, we want to prohibit edges between any pair of ghost nodes.
All of these issues can be addressed by incorporating
regularization terms proposed in \cite{Ma2018}.
Multiple connected components can be
avoided by generating graphs that have a path between
every pair of non-ghost nodes.
Using the generated tensor $\tilde{\mathbf{A}}$,
which explicitly accounts for ghost edges,
the number of paths between nodes $i$ and $j$ is given by
\begin{equation}\label{eq:num-paths}\textstyle
	\tilde{\mathbf{B}}_{ij} = I(i = j) + \sum_{k=1}^{m}
	\sum_{p = 1}^{n-1} \left( \tilde{\mathbf{A}}^p \right)_{ijk} .
\end{equation}
The regularizer comprises two terms, the first term encourages
non-ghost nodes $i$ and $j$ to be connected by a path,
and the second term that a ghost node and non-ghost node
remain disconnected:
\begin{equation}\label{eq:connectivity-penalty}\textstyle
	\frac{\mu}{n^2} \sum_{i,j}
	\left[1- \left( \tilde{\mathbf{x}}_{v_i} \right)_0 \right]
	\left[1- \left( \tilde{\mathbf{x}}_{v_j} \right)_0 \right]
	\left[1- \tilde{\mathbf{B}}_{ij} \right]
    + \frac{\mu}{n^2}
    \left( \tilde{\mathbf{x}}_{v_i} \right)_0
    \left( \tilde{\mathbf{x}}_{v_j} \right)_0
    \tilde{\mathbf{B}}_{ij} ,
\end{equation}
where $\mu$ is a hyper-parameter, and $\left( \tilde{\mathbf{x}}_{v_i} \right)_0 > 0$ if the $i$-th node is a ghost node.

To ensure atoms have valid valence, we
enforce an upper bound -- hydrogen atoms are modeled implicitly --
on the number of edges of a node, depending on its type
(e.g. four for carbon).
Let $\mathbf{u} = (u_{0}, u_1, \ldots, u_d)^\top$ be
a vector indicating the maximum capacity (number of bonding electron pairs) a node
of a given type can have, where $u_{0} = 0$ denotes the capacity
of ghost nodes. The vector $\mathbf{b} = (b_0, b_{r_1}, \ldots, b_{r_m})$
denotes the capacity for each edge type with $b_0 = 0$ representing
ghost edges.
The actual capacity of a node $v_i$ can be computed by
$
	c_{v_i} = \sum_{j \neq i} \mathbf{b}^\top \tilde{\mathbf{A}}_{ij\bullet} .
$
If $c_{v_i}$ exceeds the value in $\mathbf{u}$ corresponding to the node type
of $v_i$, the generator incurs a penalty. The valence penalty
with hyper-parameter $\nu > 0$ is defined as
\begin{equation}\label{eq:valence-penalty}%
	\frac{\nu}{n} \sum_{i=1}^n \max(0, c_{v_i} - \mathbf{u}^\top \tilde{\mathbf{x}}_{v_i} ) .
\end{equation}

\subsection{Encoder and Discriminators}
\label{sec:encoder_discriminator}

The architecture of
the encoder $g_\phi(G, \varepsilon)$, and the
three discriminators $D_\psi(G, \mathbf{z})$, $D_\eta(G_1, G_2)$,
and $D_\xi(G)$
are closely related, because they all take graphs as input.
First, we extract node-level descriptors by stacking several GCN
layers. Next, node descriptors are aggregated to obtain
a graph-level descriptor, which forms the input to a MLP.
Here, inputs are multi-graphs with $m$ edge types, which we model
by extending the Graph Isomorphism Network (GIN) architecture~\cite{Xu2019}
to multi-graphs.
Let $\mathbf{h}_{v_i}^{(l+1)}$ denote the descriptor of node $v_i$ after
the $l$-th GIN layer, with $\mathbf{h}_{v_i}^{(0)} = \mathbf{x}_{v_i}$,
then node descriptors get updated as follows:
\begin{equation}\label{eq:gin-layer}\textstyle
	\mathbf{h}_{v_i}^{(l+1)} = \tanh \left[
		\sum_{k = 1}^m \mathrm{MLP}_{r_k}^{(l)} \left(
			(1 + \epsilon^{(l)}) \mathbf{h}_{v_i}^{(l)} +
			\sum_{u \in \mathcal{N}_{r_k}(v_i)}  \mathbf{h}_u^{(l)} \right)
	\right] ,
\end{equation}
where
$\epsilon^{(l)} \in \mathbb{R}$ is a learnable weight, and
$\mathcal{N}_{r_k}(v_i) = \{ u \,|\, (u, r_k, v_i) \in \mathcal{E} \}$.
Next, graph-level node aggregation is performed.
We use skip connections \cite{Xu2018} to aggregate node-level descriptors
from all $L$ GIN layers and soft attention \cite{Li2015} to allow the network
to learn which node descriptors to use.
The graph-level descriptor $\mathbf{h}_G$ is defined as
\begin{align}\label{eq:soft-attention-pooling}
	\mathbf{h}_{v_i}^c &= \mathrm{CONCAT} (
		\mathbf{x}_{v_i}, \mathbf{h}_{v_i}^{(1)} \ldots, \mathbf{h}_{v_i}^{(L)} ) , \\
	\mathbf{h}_{v_i}^{c^\prime} &= \tanh (\mathbf{W}_1  \mathbf{h}_{v_i}^c + \mathbf{b}_1 ), &&
	\mathbf{h}_G = \textstyle
    \sum_{v \in \mathcal{V}}
	\sigma( \mathbf{W}_2 \mathbf{h}_v^{c^\prime} + \mathbf{b}_2 )
	\odot \mathbf{h}_v^{c^\prime},
\end{align}
where $\mathbf{W}$ and $\mathbf{b}$ are parameters to be learned.
Graph-level descriptors can be abstracted further by adding an
additional MLP on top, yielding $\mathbf{h}_G^\prime$.
The discriminator $D_\xi(G)$ contains a single GIN module,
$D_\eta(G_1, G_2)$ contains two GIN modules
to extract descriptors $\mathbf{h}_{G_1}^\prime$ and
$\mathbf{h}_{G_2}^\prime$, which are combined by component-wise multiplication
and fed to a 2-layer MLP.
$D_\psi(G, \mathbf{z})$ has a noise vector as second input,
which is the input to an MLP whose output is concatenated with $\mathbf{h}_G^\prime$
and linearly projected to form
$\log[ \sigma(D_\psi(G, \mathbf{z}))]$.
The encoder $g_\phi(G, \varepsilon)$ is identical
to $D_\psi(G, \mathbf{z})$,
except that its output matches the dimensionality of $\mathbf{z}$.

\section{Evaluation Metrics}\label{sec:evaluation}

To evaluate generated molecules, we adapted the metrics
proposed in~\cite{Brown2018}:
the proportion of valid, unique, and novel molecules, and a comparison
of 10 physicochemical descriptors to assess whether
new molecules have similar chemical properties as the reference data.
	\emph{Validity} is the percentage of
	molecular graphs with a single connected component and
	correct valence for all its nodes.
	\emph{Uniqueness} is the percentage of unique molecules within a
	set of $N$ randomly sampled valid molecules.
	\emph{Novelty} is the percentage of molecules not in the training
	data within a set of $N$ unique randomly sampled molecules.
Validity and novelty are percentages with respect to
the set of valid and unique molecules, which we obtain by
repeatedly sampling $N$ molecules (up to 10 times). Models that
do not generate $N$ valid/unique molecules will be penalized.
To measure to
which extent a model is able to estimate the distribution of molecules
from the training data, we first compute the distribution
of 10 chemical descriptors $d_k$ used in~\cite{Brown2018}.
\emph{Internal similarity} is a measure of diversity that
is defined as the maximum Tanimoto similarity
with respect to all other molecules in the dataset -- using the
binary Extended Connectivity Molecular Fingerprints
with diameter 4 (ECFP4)~\cite{Rogers2010}.
The remaining descriptors measure physicochemical properties of molecules
(see fig.~\ref{fig:alice-descriptors} for a full list).

For each descriptor, we compute the difference between the distribution
with respect to generated molecules~($Q$) and
the reference data~($P$) using the 1-Wasserstein distance.
This is in contrast to~\cite{Brown2018}, who use the Kullback-Leibler (KL) divergence
$D_{\text{KL}}(P\parallel Q)$.
Using the KL divergence has several drawbacks: (i) it is undefined if the
support of $P$ and $Q$ do not overlap, and (ii) it is not symmetric.
Due to the lack of symmetry,
it makes a big difference whether $Q$ has a bigger support than $P$ or the
other way around. Most importantly, an element outside of the support of
$P$ is not penalized, therefore $D_{\text{KL}}(P\parallel Q)$ will remain
unchanged whether samples of $Q$ are just outside of the support of $P$ or
far away.

The 1-Wasserstein distance, also called Earth Mover's Distance (EMD),
is a valid metric and does not have this undesirable properties.
It also offers more flexibility in how out-of-distribution samples are
penalized by choosing an appropriate ground distance (e.g. Euclidean distance for
quadratic penalty, and Manhattan distance for linear penalty).
We approximate the distribution on the reference
data and the generated data using histograms $\mathbf{h}_k^\text{ref}$ and $\mathbf{h}_k^\text{gen}$,
and define the ground distance $\mathbf{C}_{ij}(k)$ between the edges of the $i$-th
and $j$-th bin as the Euclidean distance, normalized by the minimum of the standard deviation of
descriptor $d_k$ on the reference and the generated data.
As overall measure of how well properties of generated molecules match
those of molecules in the reference data,
we compute the mEMD, defined as
\begin{align}
	\mathrm{mEMD}(\mathbf{h}_k^\text{ref}, \mathbf{h}_k^\text{gen}) &=
	{\textstyle
	\frac{1}{10} \sum_{k=1}^{10}
	\exp [- \mathrm{EMD}(\mathbf{h}_k^\text{ref}, \mathbf{h}_k^\text{gen}) ]
	,} \\
	\label{eq:EMD-metric}
	\mathrm{EMD}(\mathbf{h}_k^\text{ref}, \mathbf{h}_k^\text{gen}) &=
	{\textstyle
    \min_{\mathbf{P} \in \mathbf{U}(\mathbf{h}_k^\text{ref}, \mathbf{h}_k^\text{gen})}
	\langle \mathbf{C}(k), \mathbf{P} \rangle ,}
\end{align}
where $\mathbf{U}(\mathbf{a}, \mathbf{b})$ is the set of coupling matrices
with $\mathbf{P} \bbbone = \mathbf{a}$ and $\mathbf{P}^\top \bbbone = \mathbf{b}$ \cite{Rubner2000}.
A perfect model with $\mathrm{EMD}=0$ everywhere would obtain a mEMD score of 1.

\section{Experiments}

In our experiments, we use molecules from the
QM9 dataset \cite{Ramakrishnan2014} with at most 9 heavy atoms.
We consider $d=4$ node types (atoms C, N, O, F),
and $m=3$ edge types (single, double, and triple bonds).
After removal of molecules with non-zero formal charge, we
retained \num{131941} molecules, which we split into 80\% for training,
10\% for validation, and 10\% for testing.

We extensively compare \ourmodel\ against three state-of-the-art VAEs:
NeVAE \cite{Samanta2018} and CGVAE \cite{Liu2018} are graph-based
VAEs with validity constraints, while GrammarVAE \cite{Kusner2017} uses
the SMILES representation.
We also compare against MolGAN \cite{DeCao2018}, which is a
Wasserstein GAN without inference network, reconstruction loss,
node connectivity, or valence constraints.
Finally, we include a \emph{random graph generation model}, which only
enforces valence constraints during generation, similar to CGVAE and NeVAE,
but selects node types ($\mathbf{X}$) and edges ($\mathbf{A}$) randomly.
Note that generated graphs can have multiple connected components
if valence constraints cannot be satisfied otherwise; we consider
these to be invalid.
For NeVAE, we use our own implementation, for the remaining methods,
we use the authors' publicly available code.
Further details are described in supplement \ref{sec:supp_implementation}.

\subsection{Latent Space}

\newsavebox{\tempbox}
\begin{figure}[tb]
    \centering
    \sbox{\tempbox}
    {\includegraphics[height=3.6cm]{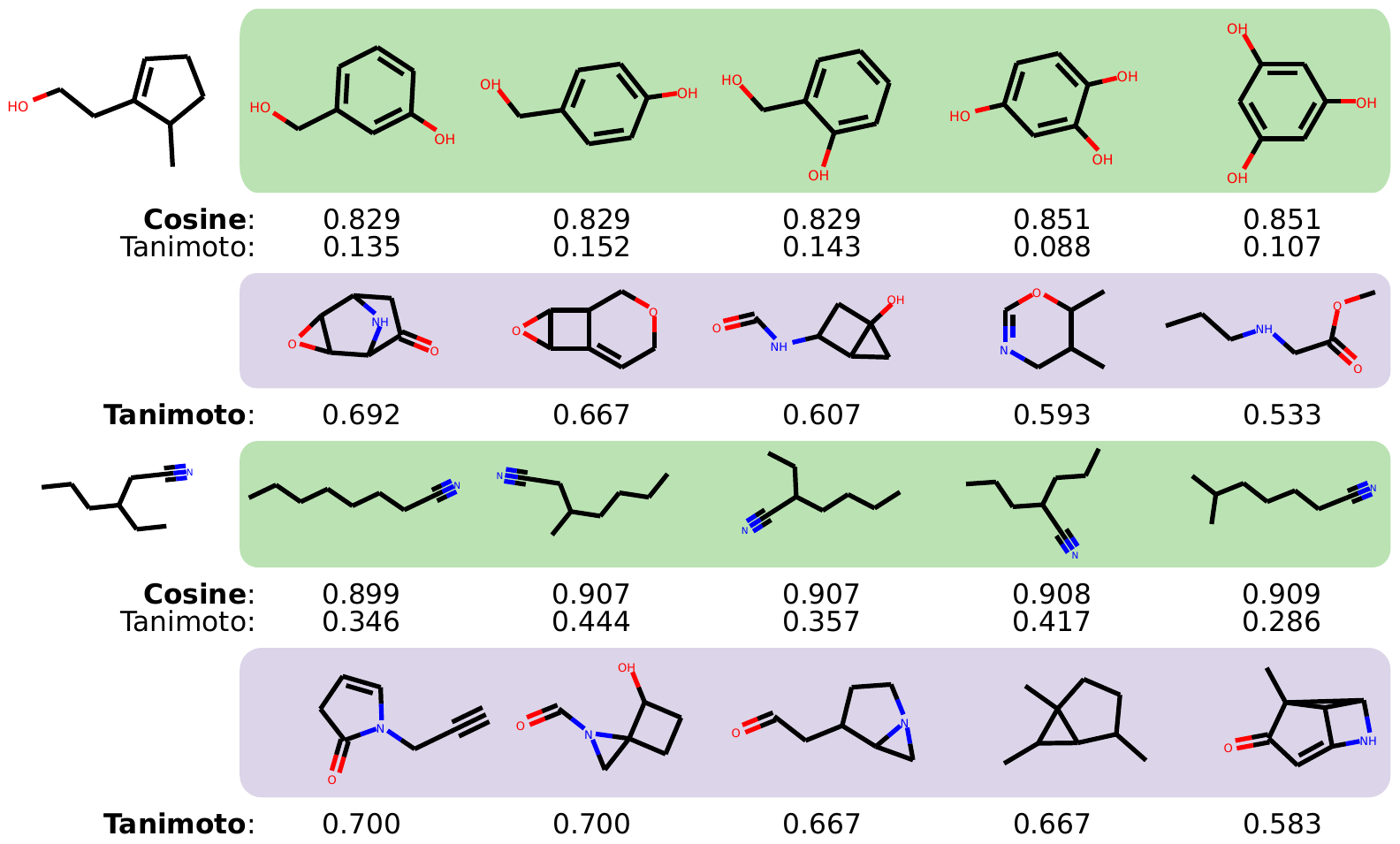}}
    \begin{minipage}[t][][b]{0.49\textwidth}%
        {\footnotesize(a)}\\
        \usebox{\tempbox}%
    \end{minipage}\hfill%
    \begin{minipage}[t][][b]{0.49\textwidth}%
        {\footnotesize(b)}\\
        \vbox to \ht\tempbox{%
            \vfil
            \setlength{\tabcolsep}{2pt}\tiny%
            \begin{tabular}{lrr}
                \toprule
                {} & \multicolumn{1}{c}{Reconstructed (\%)} & \multicolumn{1}{c}{Generated (\%)} \\
                \midrule
                Ghost-node bond       &             0 ~(0.00) &         2 ~(0.02) \\
                Valency               &            17 ~(0.13) &        36 ~(0.28) \\
                Split graph           &           296 ~(2.31) &       686 ~(5.36) \\
                Valid                 &   \num{12487} (97.55) & \num{12076} (94.34) \\
                \bottomrule
                \end{tabular}%
            \vfil}%
    \end{minipage}%
    \caption{\label{fig:nearest-neighbors}%
        (a) Nearest neighbors with respect to the molecule in the first
        column in embedding space (green rows)
        and according to Tanimoto similarity (purple rows).
        (b) Frequency of errors of \ourmodel\ on test set.}
\end{figure}

In the first experiment, we investigate properties of the encoder and the
associated latent space.
We project molecules of the
test set into the latent space, and perform
a $k$ nearest neighbor search to find the
closest latent representation of a molecule
in the training set in terms of cosine distance.
We compare results against the nearest neighbors
by Tanimoto similarity of ECFP4 fingerprints \cite{Rogers2010}.
Figure~\ref{fig:nearest-neighbors}a shows
that the two approaches lead to quite
different sets of nearest neighbors.
Nearest neighbors based on molecules'
latent representation usually differ by
small substructures. For instance,
the five nearest neighbor in the first row
of fig.~\ref{fig:nearest-neighbors}a
differ by the location of side chains
around a shared ring structure.
On the other hand, the topology of nearest neighbors by
Tanimoto similarity (second row) differs
considerably from the query.
Moreover, all but one nearest neighbor contain
nitrogen atoms, which are absent from the query.
In the second example (last two rows),
the nearest neighbors in latent space are all
linear structures with one triple bond and nitrogen,
whereas all nearest neighbors by Tanimoto similarity
contain ring structures and only one
contains a triple bond.
Additional experiments with respect to interpolation in the latent space
can be found in supplement \ref{sec:supp_latent_space}.

\subsection{Molecule Generation}

Next, we evaluate the quality of
generated molecules.
We generate molecules from $N=\num{10000}$ latent vectors,
sampled from a unit sphere Gaussian
and employ the metrics described in section~\ref{sec:evaluation}.
Note that previous work defined uniqueness and novelty as the percentage
with respect to all valid molecules rather than $N$, which is hard to interpret,
because models with low validity would have high novelty.
Hence, percentages reported here are considerably lower.

\begin{figure}[tb]
    \centering
    \begin{minipage}[t][][b]{0.03\textwidth}{\scriptsize(a)}\end{minipage}%
    \begin{minipage}[t][][b]{0.57\textwidth}%
    {\includegraphics[height=3.5cm,trim={0 0 3.6cm 0},clip]{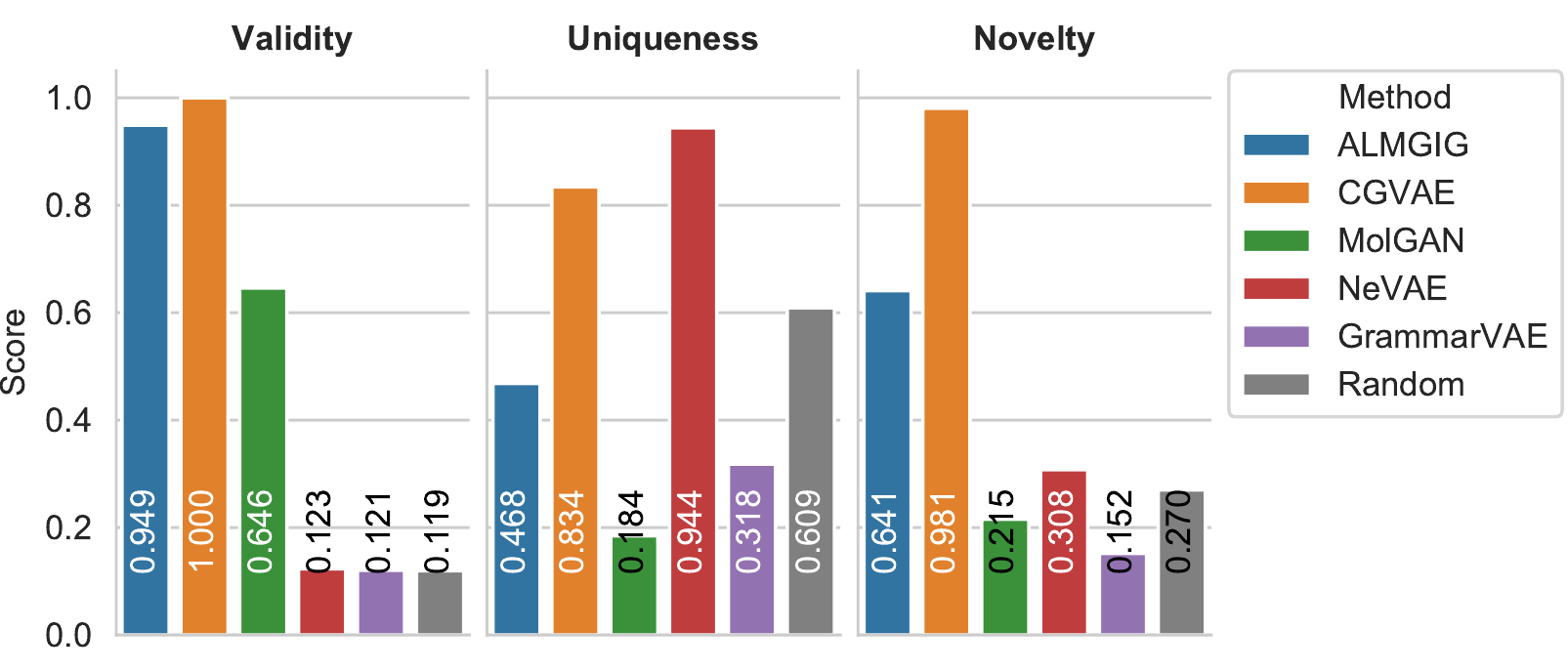}}%
    \end{minipage}%
    \begin{minipage}[t][][b]{0.03\textwidth}{\scriptsize(b)}\end{minipage}%
    \begin{minipage}[t][][b]{0.37\textwidth}%
    {\includegraphics[height=3.5cm]{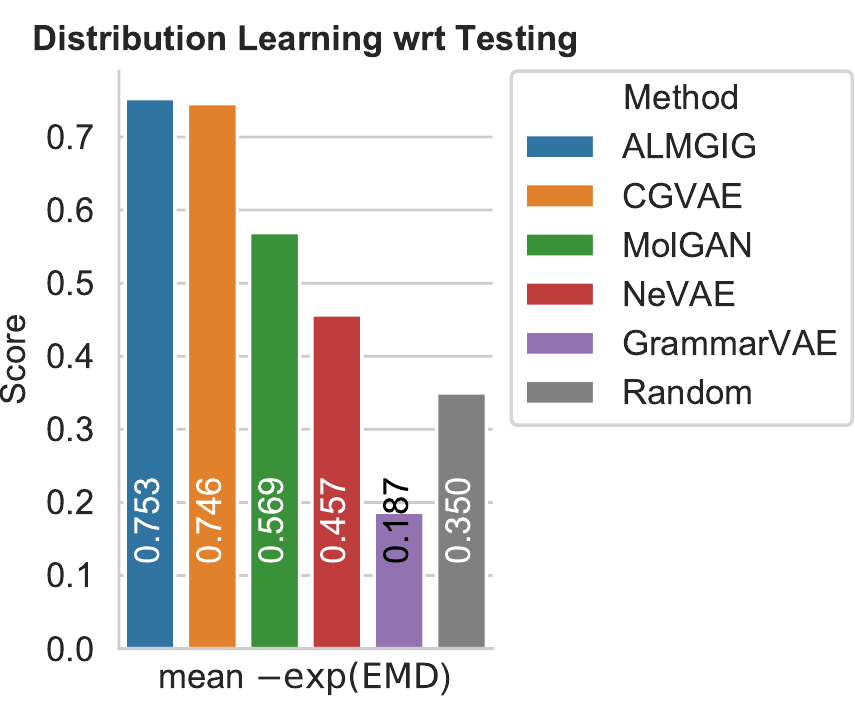}}%
    \end{minipage}
    \caption{\label{fig:method-compare}%
        (a) Overview of simple molecule generation statistics.
        (b) Comparison with respect to the proposed mEMD evaluation metric for
        distribution learning. See figs.~\ref{fig:alice-descriptors} and
        \ref{fig:more-descriptors} for differences between individual distributions.}%
\end{figure}

Figure~\ref{fig:method-compare}a shows that \ourmodel\ generates
molecules with high validity (94.9\%) and is only outperformed by CGVAE,
which by design is constrained to only generate valid molecules, but
has ten times more parameters than \ourmodel\ (1.1M vs. 13M).
Moreover, \ourmodel\ ranks second in novelty and third in uniqueness (excluding random);
we will investigate the reason for this difference in detail in the next section
on distribution learning.
Graph-based NeVAE always generates molecules with correct
valence, but often (88.2\%) generates graphs with multiple connected components,
which we regard as invalid.
Its set of valid generated molecules has the highest uniqueness.
We can also observe that graph-based models outperform the
SMILES-based GrammarVAE, which is prone to generate
invalid SMILES representation of molecules, which is
a known problem%
~\cite{Blaschke2018,Guimaraes2017,Lim2018,Putin2018,Segler2018}.
MolGAN is a regular GAN without inference network.
It is struggling to generate molecules
with valid valence and only learned one particular mode of the distribution,
which we will discuss in more detail in the next section.
It is inferior to \ourmodel\ in all categories, in particular with respect
to novelty and uniqueness of generated molecules.

Next, we inspect the reason for generated molecules being invalid
to assess whether imposed node connectivity and
valence constraints are effective.
Molecules can be generated by (a) reconstructing
the latent representation of a real molecule in
the	test set, or (b) by decoding a random latent representation
draw from an isotropic Gaussian (see fig.~\ref{fig:model-overview}).
Figure~\ref{fig:nearest-neighbors}b reveals that most errors are due to
multiple disconnected graphs being generated (2-5\%).
While individual components do represent valid molecules, we
treat them as erroneous molecular graphs.
In these instances, 98-99\% of graphs have 2 connected components
and the remainder has 3.
Less than 0.3\% of molecules have atoms with improper valence
and less than 0.1\% of graphs have atomic bonds between
ghost nodes.
Therefore, we conclude that the constraints are highly
effective.

Finally, we turn to the random graph generation model.
It achieves a relatively high uniqueness
of 60.9\%, which ranks third, and has a higher novelty
than MolGAN and GrammarVAE.
Many generated graphs have multiple connected components,
which yields a low validity.
The fact that the random model cannot be clearly distinguished from
trained models, indicates that validity, uniqueness, and
novelty do not accurately capture what we are really interested in:
\emph{Can we generate chemically meaningful molecules with similar properties as in the training data?}
We will investigate this question next.

\subsection{Distribution Learning}

While the simple overall statistics in the previous section
can be useful rough indicators, they ignore the physicochemical
properties of generated molecules and do not capture to
which extent a model is able to estimate the distribution of molecules
from the training data.
Therefore, we compare the distribution of 10 chemical descriptors
in terms of EMD \eqref{eq:EMD-metric}.
Distributions of individual descriptors are depicted in
fig.~\ref{fig:alice-descriptors} and fig.~\ref{fig:more-descriptors} of the supplement.

First of all, we want to highlight that using the proposed mEMD score, we can easily
identify the random model (see fig.~\ref{fig:method-compare}b),
which is not obvious from the simple
summary statistics in fig.~\ref{fig:method-compare}a.
In particular, from fig.~\ref{fig:method-compare}a we could have concluded that
NeVAE is only marginally better than the random model.
The proposed scheme clearly demonstrates that NeVAE is superior to
the random model (overall score 0.457 vs 0.350).
Considering differences between individual descriptors
reveals that randomly generated molecules are not meaningful
due to higher number of hydrogen acceptors, molecular weight,
molecular complexity, and polar surface area
(see fig.~\ref{fig:more-descriptors}d).

\begin{figure*}[tb]
    \centering
    \begin{minipage}[t][][b]{0.03\textwidth}{\scriptsize(a)}\end{minipage}%
    \begin{minipage}[t][][b]{0.85\textwidth}%
        \includegraphics[width=\linewidth]{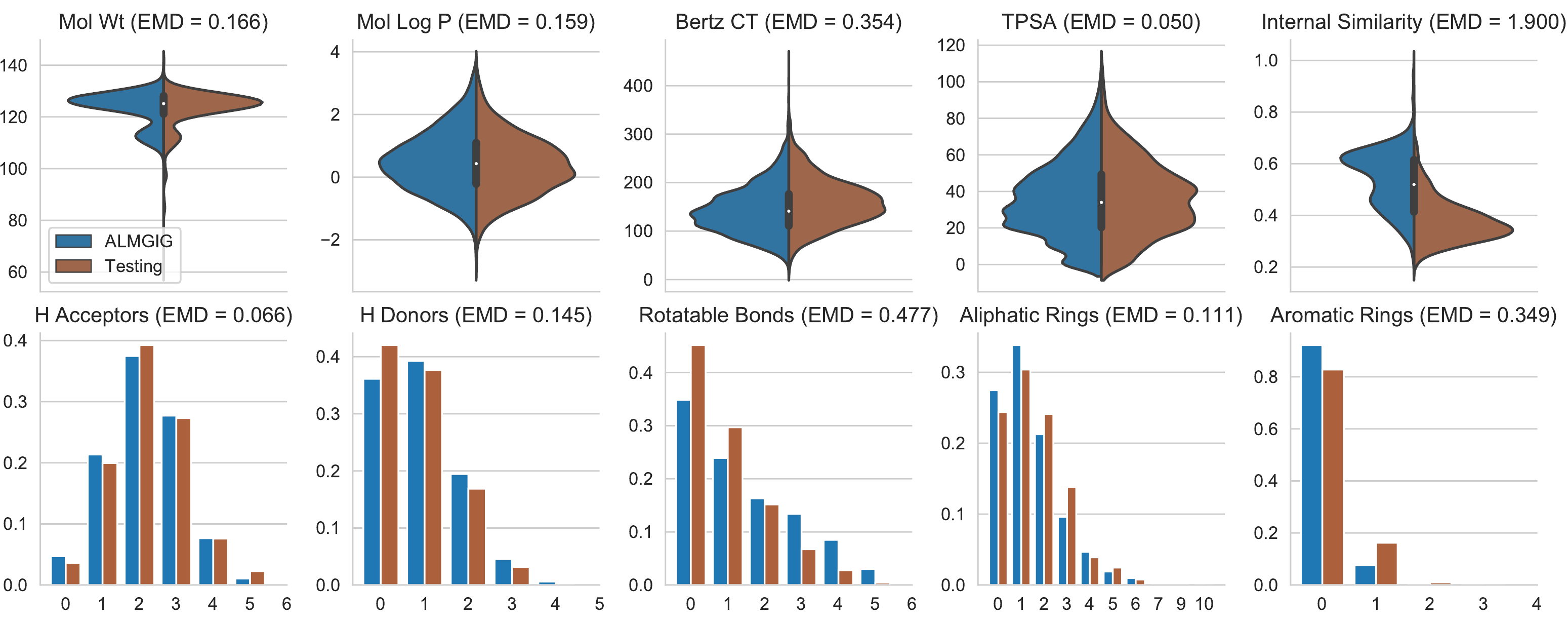}%
    \end{minipage}\vfill%
    \begin{minipage}[t][][b]{0.03\textwidth}{\scriptsize(b)}\end{minipage}%
    \begin{minipage}[t][][b]{0.85\textwidth}%
        \includegraphics[width=\linewidth]{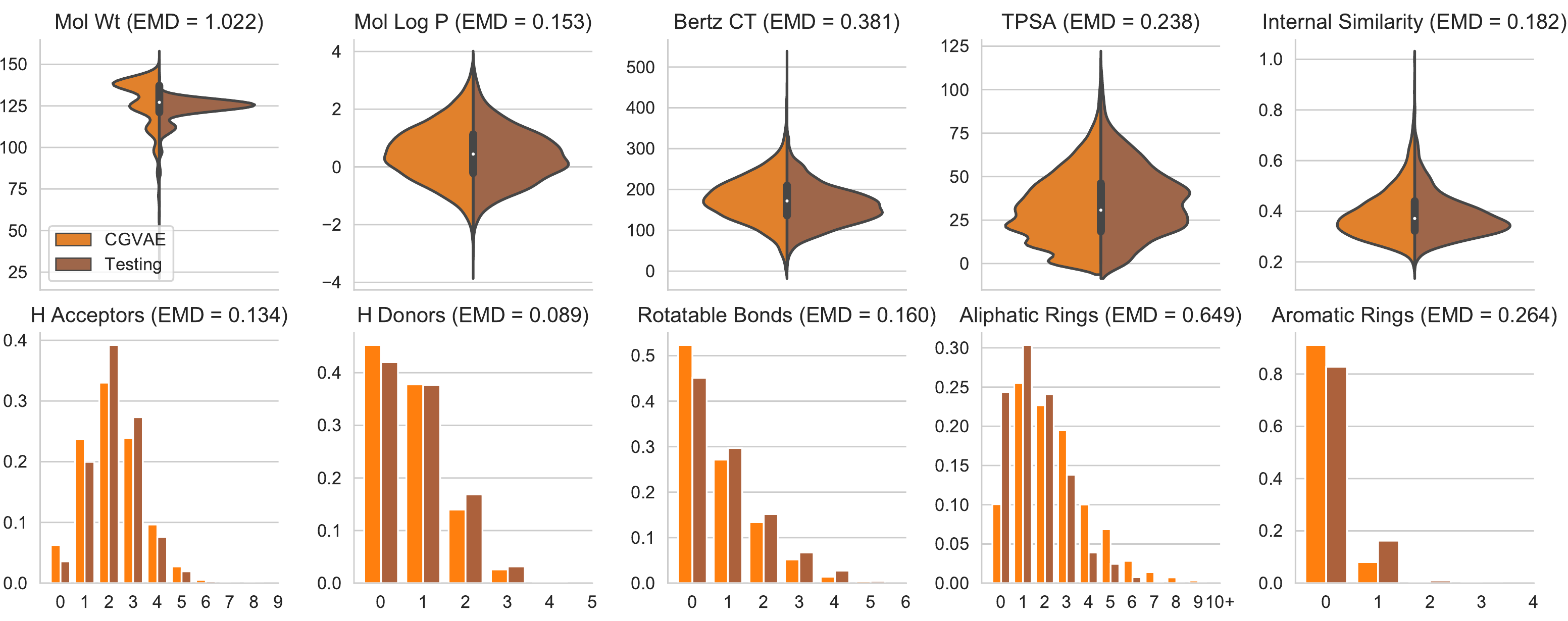}%
    \end{minipage}
    \caption{\label{fig:alice-descriptors}%
        Distribution of chemical properties of the test data
        and of 10,000 unique molecules generated by (a) \ourmodel,
        and (b) CGVAE \cite{Liu2018}.
        Mol Wt: molecular weight. Mol Log P: water-octanol partition coefficient.
        Bertz CT: molecular complexity index. TPSA: molecular polar surface area.
        H Acceptors (Donors): Number of hydrogen acceptors (donors).}
\end{figure*}

\ourmodel\ achieves the best overall mEMD score, with
molecular weight and polar surface area
best matching that of the test data by a large margin (EMD = 0.17 and 0.05,
see fig.~\ref{fig:alice-descriptors}a).
The highest difference is due to the internal similarity of generated molecules
(EMD = 1.90).
Among all VAEs, CGVAE is performing best
(see fig.~\ref{fig:alice-descriptors}b).
However, it notably produces big molecules with larger weight (EMD = 1.02)
and lower polar surface area (EMD = 0.238).
The former explains its high novelty value in fig.~\ref{fig:method-compare}a:
by creating bigger molecules than in the data,
a high percentage of generated molecules is novel and unique.
Molecules generated by NeVAE (see fig.~\ref{fig:more-descriptors}a) have issues similar to CGVAE:
they have larger weight (EMD = 1.72) and higher polar surface area (EMD = 0.839),
which benefits uniqueness.
In contrast to \ourmodel\ and CGVAE, it also struggles to generate molecules with aromatic rings (EMD = 0.849).
GrammarVAE in fig.~\ref{fig:more-descriptors}b is far from capturing the data distribution, because it tends
to generate long SMILES strings corresponding to heavy molecules
(EMD = 5.66 for molecular weight).
Molecules produced by MolGAN are characterized by a striking increase
in internal similarity without any overlap with the reference distribution
(EMD = 6.41;  fig.~\ref{fig:more-descriptors}c).
This result is an example where comparison by KL divergence would be undefined.
An interesting detail can be derived from the distribution of molecular weight:
MolGAN has problems generating molecules
with intermediate to low molecular weight ($<115$\,g/mol), which
highlights a common problem with GANs, where only one mode of the distribution
is learned by the model (mode collapse).
In contrast, \ourmodel\ can capture this mode, but still misses the smaller
mode with molecular weight below 100\,g/mol (see fig.~\ref{fig:alice-descriptors}a).
This mode contains only \num{2344} molecules (2.2\%), which
makes it challenging -- for any model -- to capture.
This demonstrates that our proposed Wasserstein-based evaluation
can reveal valuable insights that would have been missed
when solely relying on validity, novelty, and uniqueness.

\subsection{Ablation Study}

\begin{table}[tb]
    \centering
    \caption{\label{tab:ablation-study}%
        Configurations evaluated in our ablation study.
        SC: Skip-connections.}
    \begin{scriptsize}
    \begin{tabular}{lcc|ccc|ccc}
        \toprule
        {} & \multicolumn{2}{c|}{Penalties}
        & \multicolumn{3}{c|}{Discriminators}
        & \multicolumn{3}{c}{Architecture} \\
        {} & Conn. & Valence & Unary & Joint & Cycle & GIN SC & Gen. SC & Attention \\
        \midrule
        \ourmodel\ & \cmark & \cmark & \cmark & \cmark & \cmark & \cmark & \cmark & \cmark \\
        No Connectivity& \xmark & \cmark & \cmark & \cmark & \cmark & \cmark & \cmark & \cmark \\
        No Valence& \cmark & \xmark & \cmark & \cmark & \cmark & \cmark & \cmark & \cmark \\
        No Conn+Valence& \xmark & \xmark & \cmark & \cmark & \cmark & \cmark & \cmark & \cmark \\
        No GIN SC& \cmark & \cmark & \cmark & \cmark & \cmark & \xmark & \cmark & \cmark \\
        No Generator SC& \cmark & \cmark & \cmark & \cmark & \cmark & \cmark & \xmark & \cmark \\
        No Attention & \cmark & \cmark & \cmark & \cmark & \cmark & \cmark & \cmark & \xmark \\
        ALICE & \cmark & \cmark & \xmark & \cmark & \cmark & \cmark & \cmark & \cmark \\
        ALI & \cmark & \cmark & \xmark & \cmark & \xmark & \cmark & \cmark & \cmark \\
        (W)GAN & \cmark & \cmark & \cmark & \xmark & \xmark & \cmark & \cmark & \cmark \\
        \bottomrule
    \end{tabular}
    \end{scriptsize}
\end{table}

Next, we evaluate a number of modeling choices by
conducting an extensive ablation study (see table~\ref{tab:ablation-study}).
We first evaluate the impact of
the connectivity \eqref{eq:connectivity-penalty}
and valence penalty \eqref{eq:valence-penalty} during training.
Next, we evaluate architectural choices, namely
(i) skip connections in the GIN of encoder and discriminators,
(ii) skip connections in the generator,
and
(iii) soft attention in the graph pooling layer \eqref{eq:soft-attention-pooling}.
Finally, we use the proposed architecture and compared different
adversarial learning schemes:
ALI~\cite{Dumoulin2017}, ALICE~\cite{Li2017}, and a traditional
WGAN without an encoder network $g_\phi(G, \varepsilon)$~\cite{Arjovsky2017,Goodfellow2014}.
Note that WGAN is an extension of MolGAN~\cite{DeCao2018} with
connectivity and valence penalties, and
GIN-based architecture.

Our results are summarized in fig.~\ref{fig:ablation-study-results}
(details in fig.~\ref{fig:descriptors-ablation} of the supplement).
As expected, removing valence and connectivity penalties
lowers the validity, whereas the drop when only removing
the valence penalty is surprisingly small, but does lower
the uniqueness considerably.
Regarding our architectural choices, results demonstrate that
only the proposed architecture of \ourmodel\ is able to generate
a sufficient number of molecules with aromatic rings (EMD = 0.35),
when removing the attention mechanism (EMD = 0.96),
skip connections from the GIN (EMD = 0.74), or
from the generator (EMD = 1.26) the EMD increases at least
two-fold.
When comparing alternative adversarial learning schemes,
we observe that several configurations resulted in mode collapse.
It is most obvious for ALI, GAN, and WGAN, which can only generate
a few molecules and thus have very low uniqueness and novelty.
In particular, ALI and WGAN are unable to generate molecules with
aromatic rings.
ALICE is able to generate more diverse molecules, but is capturing
only a single mode of the distribution: molecules with low to medium
weight are absent.
This demonstrates that including a unary discriminator, as in \ourmodel,
is vital to capture the full distribution over chemical structures.
Finally, it is noteworthy that WGAN,
using our proposed combination of GIN architecture and penalties,
outperforms MolGAN~\cite{DeCao2018}. This demonstrates that our
proposed architecture can already improve existing methods for
molecule generation considerably.
When also extending the adversarial learning framework, the
results demonstrate that \ourmodel\ can capture the underlying
distribution of molecules more accurately than any competing method.

\begin{figure}[tb]
    \centering
    \begin{minipage}[t][][b]{0.03\textwidth}{\scriptsize(a)}\end{minipage}%
    \begin{minipage}[t][][b]{0.97\textwidth}%
    {\includegraphics[height=3.5cm,trim={0 0 5.3cm 0},clip]{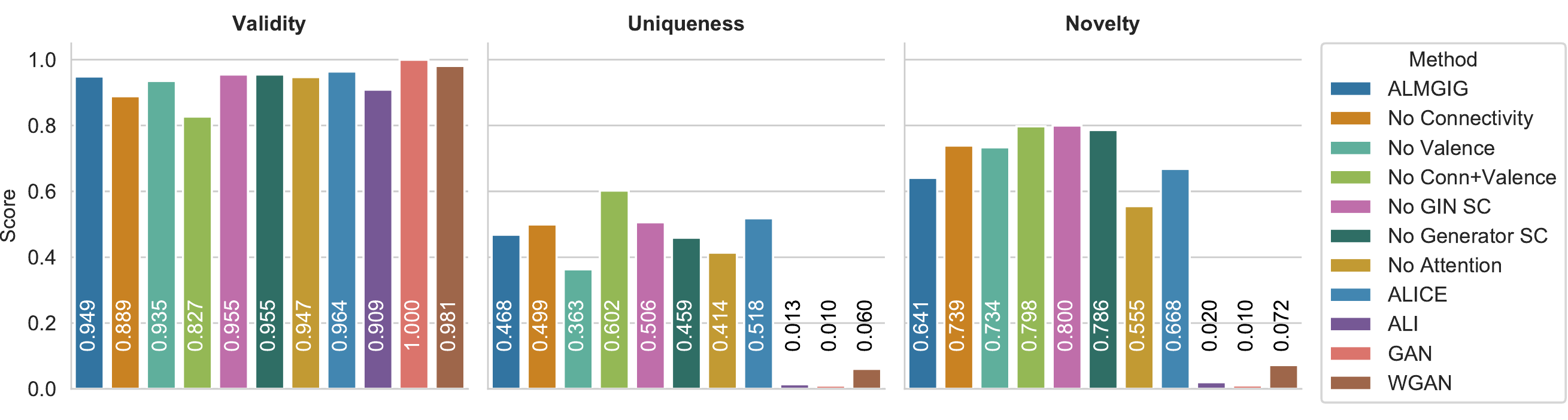}}%
    \end{minipage}
    \begin{minipage}[t][][b]{0.03\textwidth}{\scriptsize(b)}\end{minipage}%
    \begin{minipage}[t][][b]{0.97\textwidth}%
    {\includegraphics[height=3.5cm]{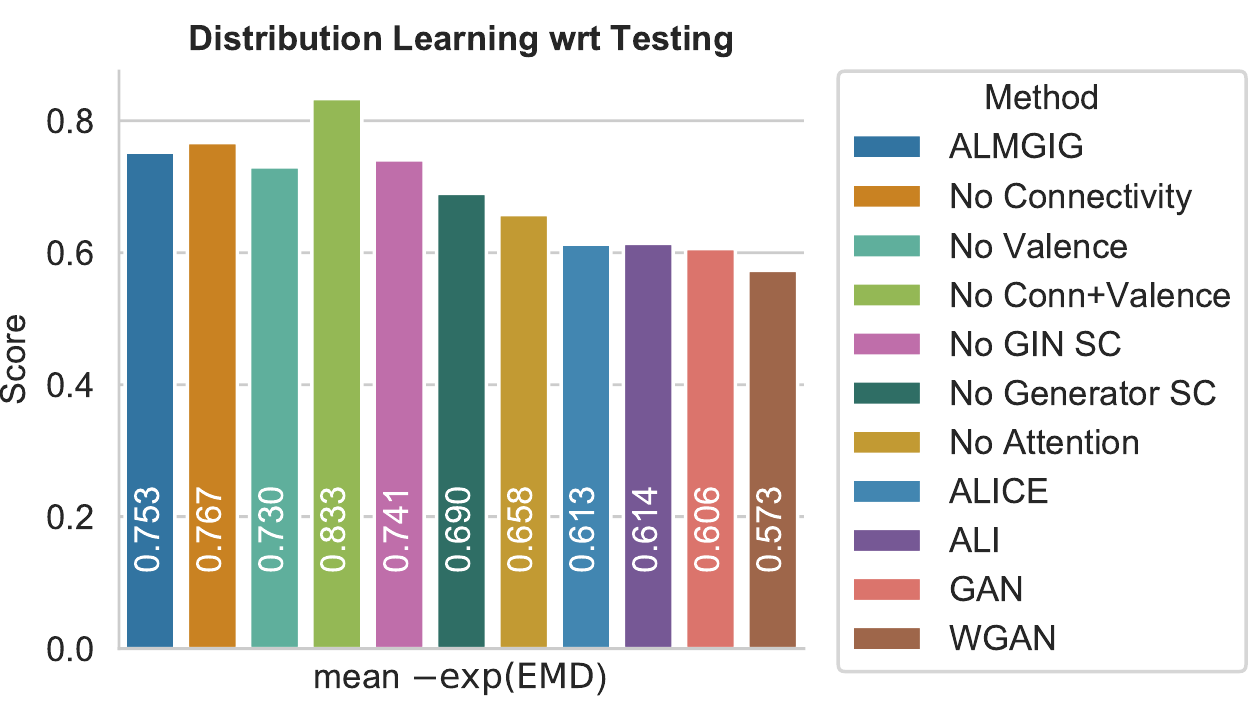}}%
    \end{minipage}
    \caption{\label{fig:ablation-study-results}%
        Results of ablation study.
        (a) Simple molecule generation statistics.
        (b) Comparison with respect to the proposed mEMD evaluation metric for
        distribution learning.}%
\end{figure}

\section{Conclusion}

We formulated generation and inference of molecular graphs
as a likelihood-free adversarial learning task.
Compared to previous work, it allows training without
explicitly computing a reconstructing loss, which would require
solving an expensive graph isomorphism problem.
Moreover, we argued that the common validation metrics validity, novelty,
and uniqueness are insufficient to properly assess the performance of
algorithms for molecule generation, because they ignore
physicochemical properties of generated molecules.
Instead, we proposed to compute the 1-Wasserstein distance
between distributions of physicochemical properties of molecules.
We showed that the proposed
adversarial learning framework for molecular graph inference and generation, \ourmodel,
allows efficiently
exploring the space of molecules via
molecules' continuous latent representation,
and that it more accurately represents the distribution over the space
of molecules than previous methods.

\subsubsection*{Acknowledgements}
This research was partially supported by the Bavarian State Ministry of Education,
Science and the Arts in the framework of the Centre Digitisation.Bavaria (ZD.B),
and the Federal Ministry of Education and Research in the call for Computational Life Sciences (DeepMentia, 031L0200A).
We gratefully acknowledge the support of NVIDIA Corporation with the donation of
the Quadro P6000 GPU used for this research.

\bibliography{references}
\bibliographystyle{splncs04}

\appendix
\counterwithin{figure}{section}
\counterwithin{table}{section}

\section{Additional Results}
\subsection{Latent Space Interpolation}\label{sec:supp_latent_space}

\begin{figure}
    \centering
    \includegraphics[width=.8\linewidth]{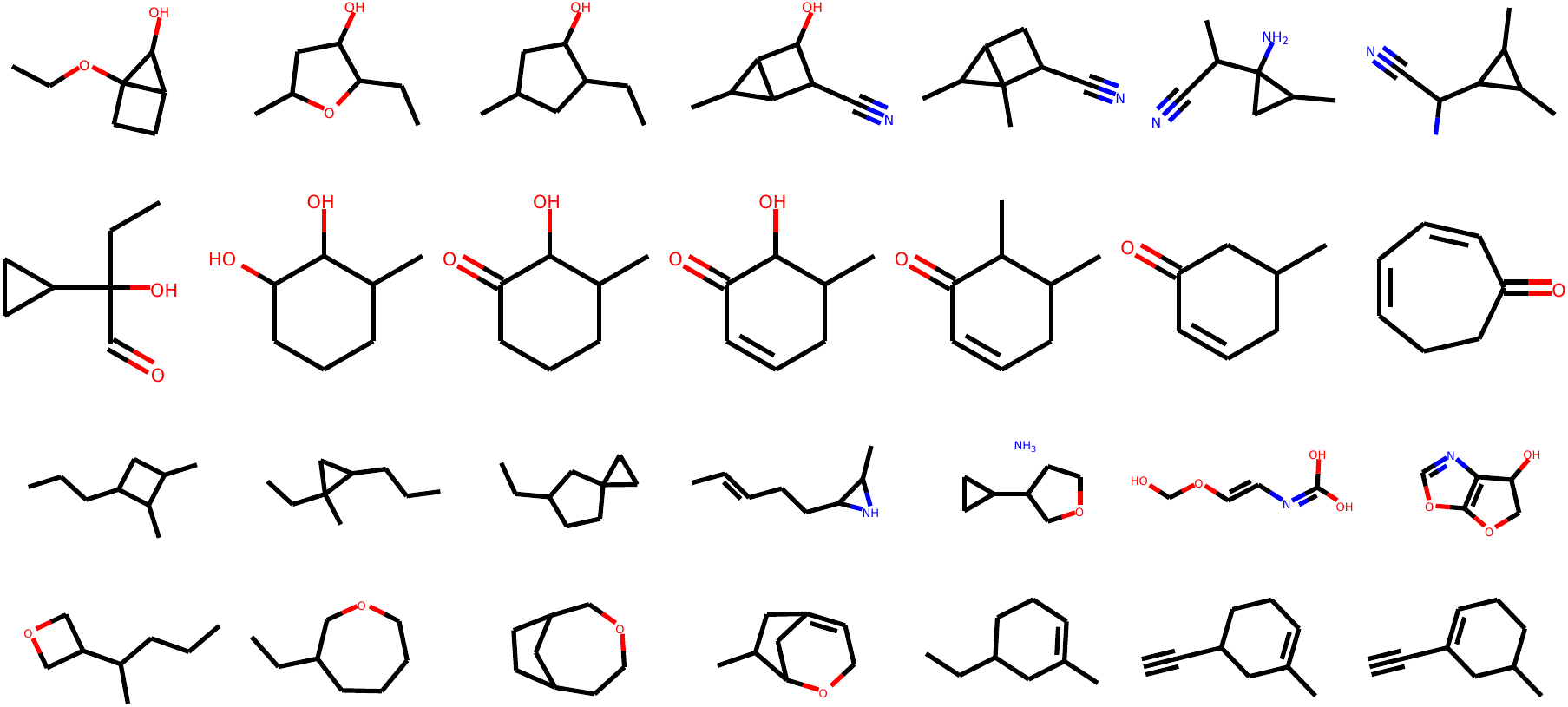}
    \caption{\label{fig:interpolation-between-real}%
        Each row is a one-dimensional interpolation in latent space.
        The left and right most molecule are from the QM9 data, the remaining
        molecules lie on the line from the latent representation of the left most
        to the right most molecule.}
\end{figure}

In this experiment, we interpolate between two molecules by
computing their respective latent representation $\tilde{\mathbf{z}}_A$ and $\tilde{\mathbf{z}}_B$
and reconstructing molecules from latent codes
on the line between $\tilde{\mathbf{z}}_A$ and $\tilde{\mathbf{z}}_B$.
The results are depicted in fig.~\ref{fig:interpolation-between-real}.
Interpolation between molecules appears smooth, but we can also
see examples of latent representations
corresponding to graphs with multiple connected
components (second to last row), which we excluded
from our analyses above, but included here
for illustration.

\clearpage
\subsection{Distribution Learning}
\begin{figure}[h]
    \centering
    \caption{\label{fig:more-descriptors}%
        Distribution of 10 descriptors of molecules in the test data
        and of 10,000 unique molecules generated by (a) NeVAE \cite{Samanta2018},
        (b) GrammarVAE \cite{Kusner2017}, (c) MolGAN \cite{DeCao2018},
        and (d) randomly.}
    \begin{minipage}[t][][b]{0.03\textwidth}{\scriptsize(a)}\end{minipage}%
    \begin{minipage}[t][][b]{0.85\textwidth}%
        \includegraphics[width=\linewidth]{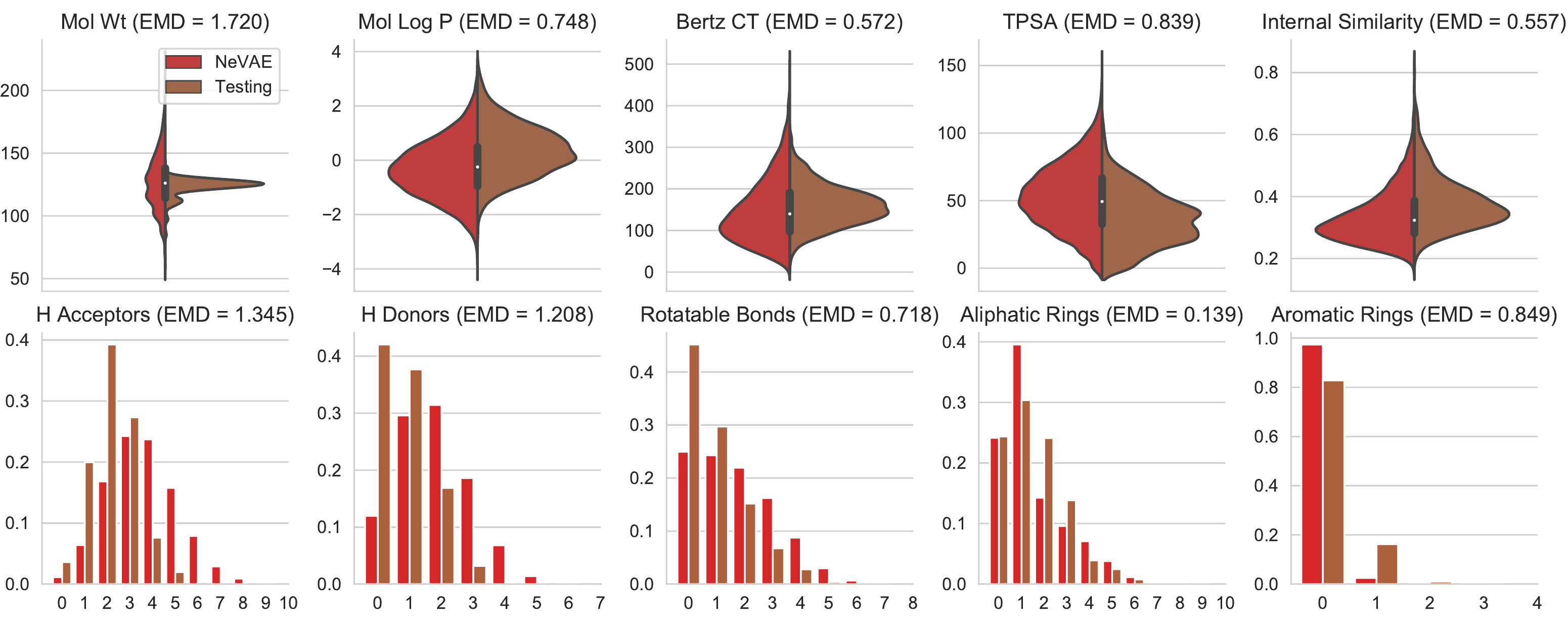}%
    \end{minipage}\vfill%
    \begin{minipage}[t][][b]{0.03\textwidth}{\scriptsize(b)}\end{minipage}%
    \begin{minipage}[t][][b]{0.85\textwidth}%
        \includegraphics[width=\linewidth]{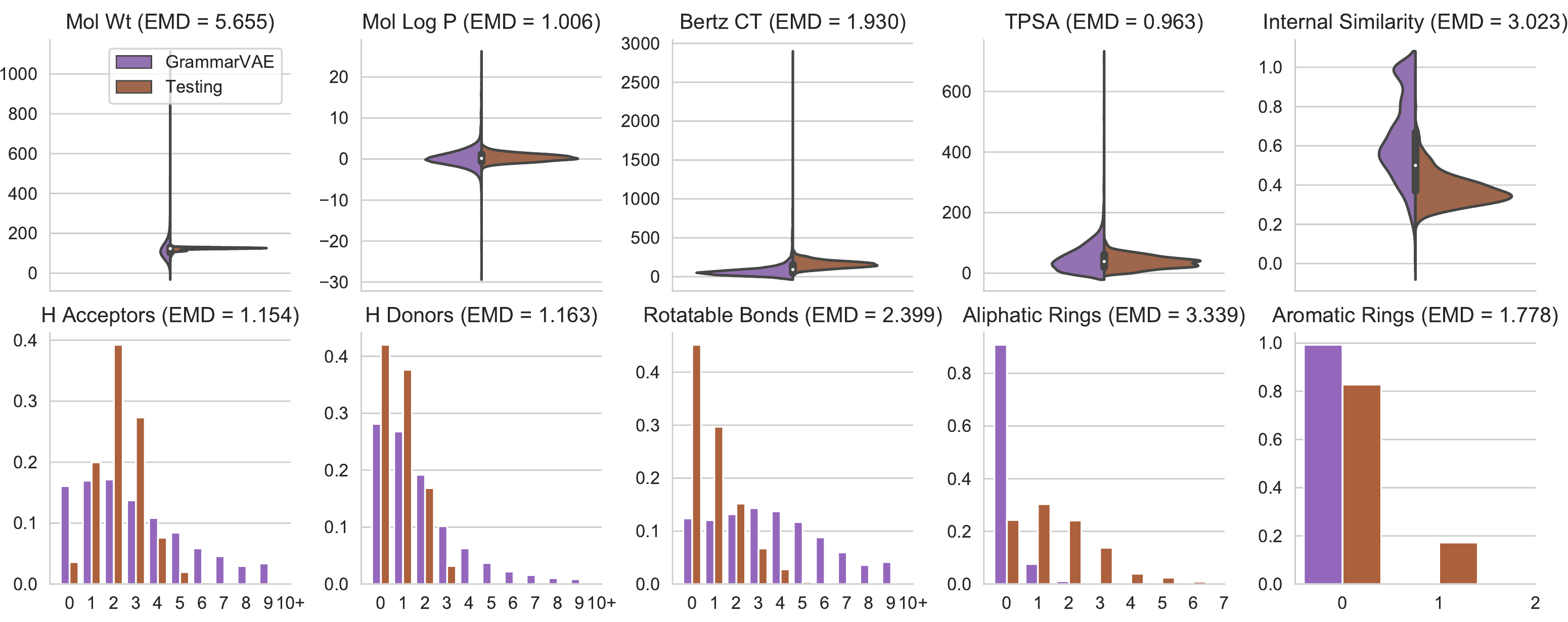}%
    \end{minipage}\vfill%
    \begin{minipage}[t][][b]{0.03\textwidth}{\scriptsize(c)}\end{minipage}%
    \begin{minipage}[t][][b]{0.85\textwidth}%
        \includegraphics[width=\linewidth]{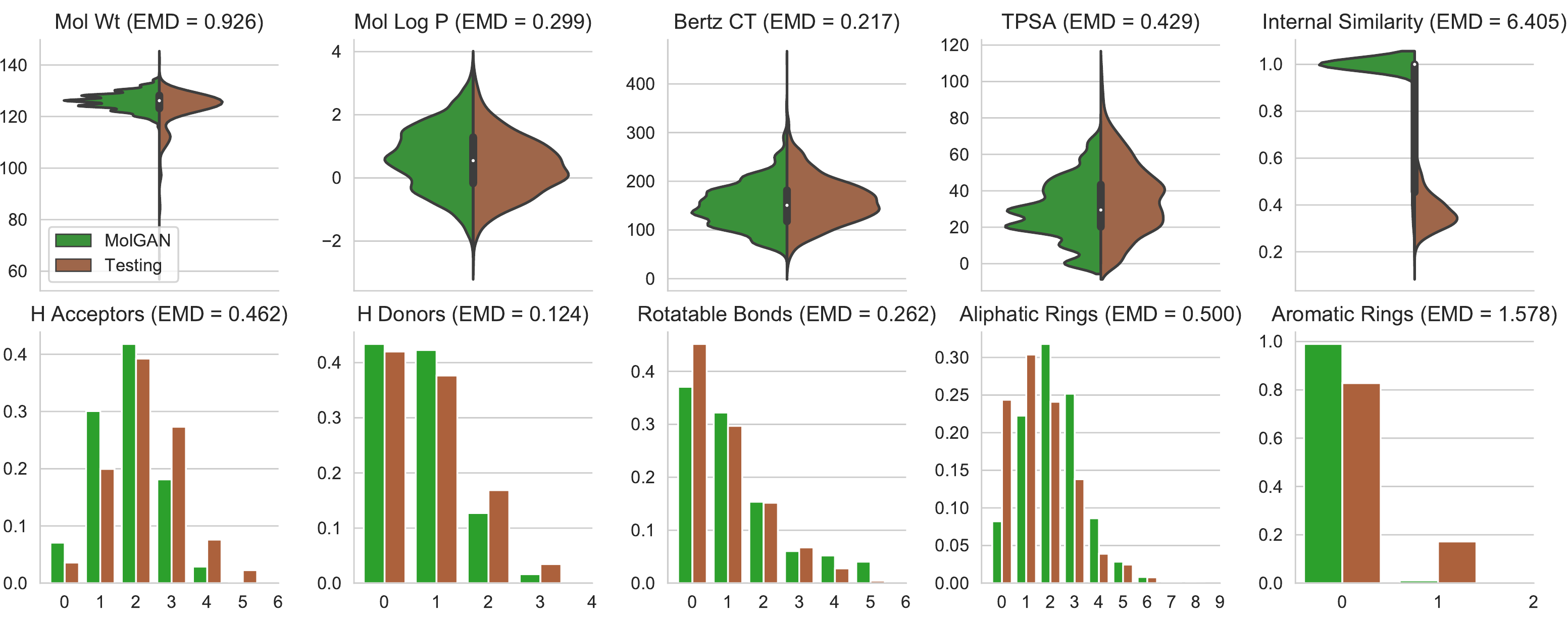}%
    \end{minipage}
\end{figure}
\begin{figure}[h]
    \centering
    \begin{minipage}[t][][b]{0.03\textwidth}{\scriptsize(d)}\end{minipage}%
    \begin{minipage}[t][][b]{0.85\textwidth}%
    \includegraphics[width=\linewidth]{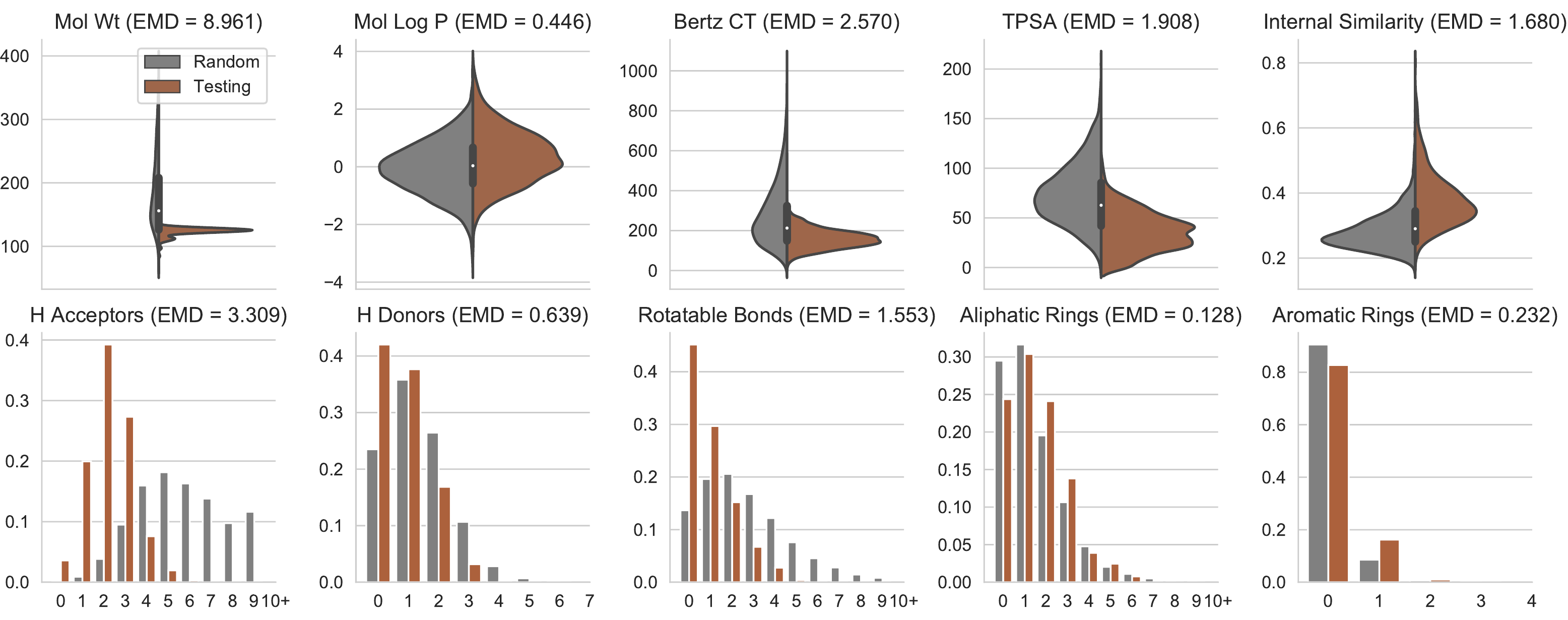}%
    \end{minipage}
\end{figure}
\begin{figure}[h]
    \centering
    \caption{\label{fig:descriptors-ablation}%
    Distribution of 10 descriptors of molecules in the test data
    and of 10,000 unique generated molecules used in ablation study.}
    \begin{minipage}[t][][b]{0.03\textwidth}{\scriptsize(a)}\end{minipage}%
    \begin{minipage}[t][][b]{0.85\textwidth}%
        \includegraphics[width=\linewidth]{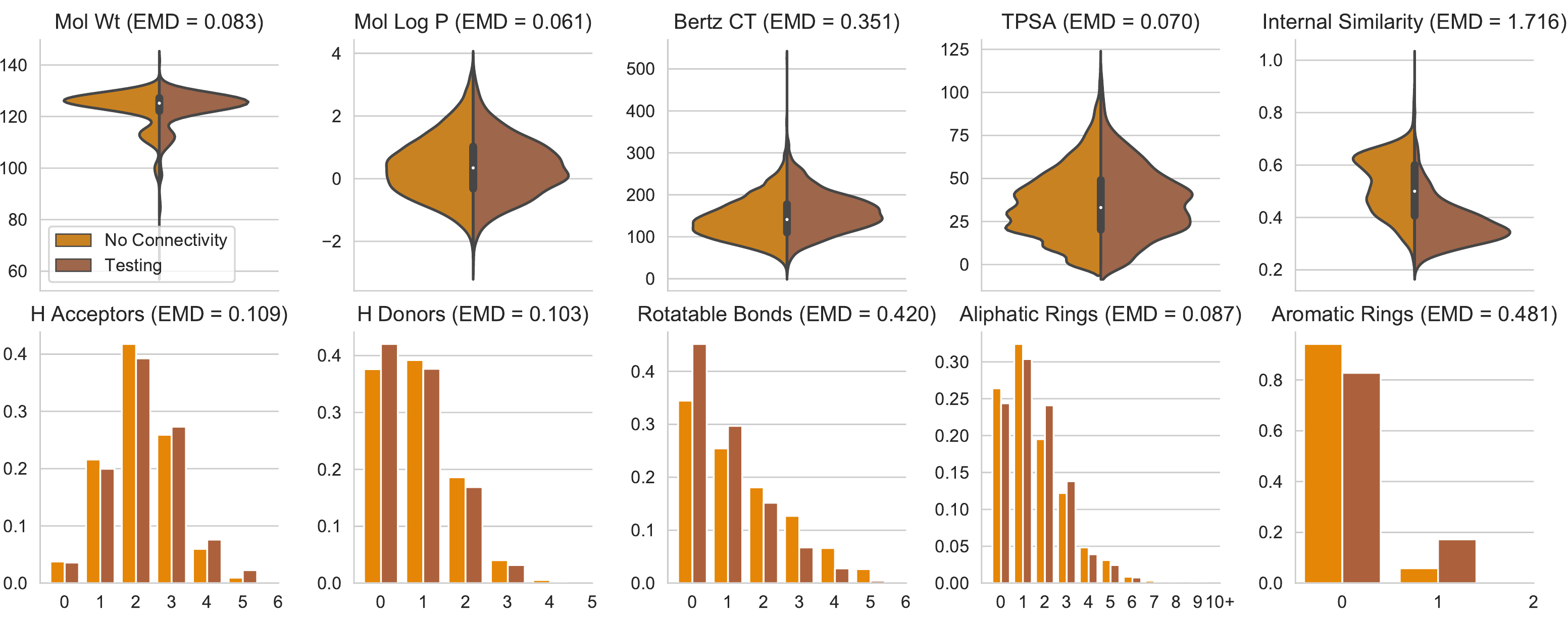}%
    \end{minipage}\vfill%
    \begin{minipage}[t][][b]{0.03\textwidth}{\scriptsize(b)}\end{minipage}%
    \begin{minipage}[t][][b]{0.85\textwidth}%
        \includegraphics[width=\linewidth]{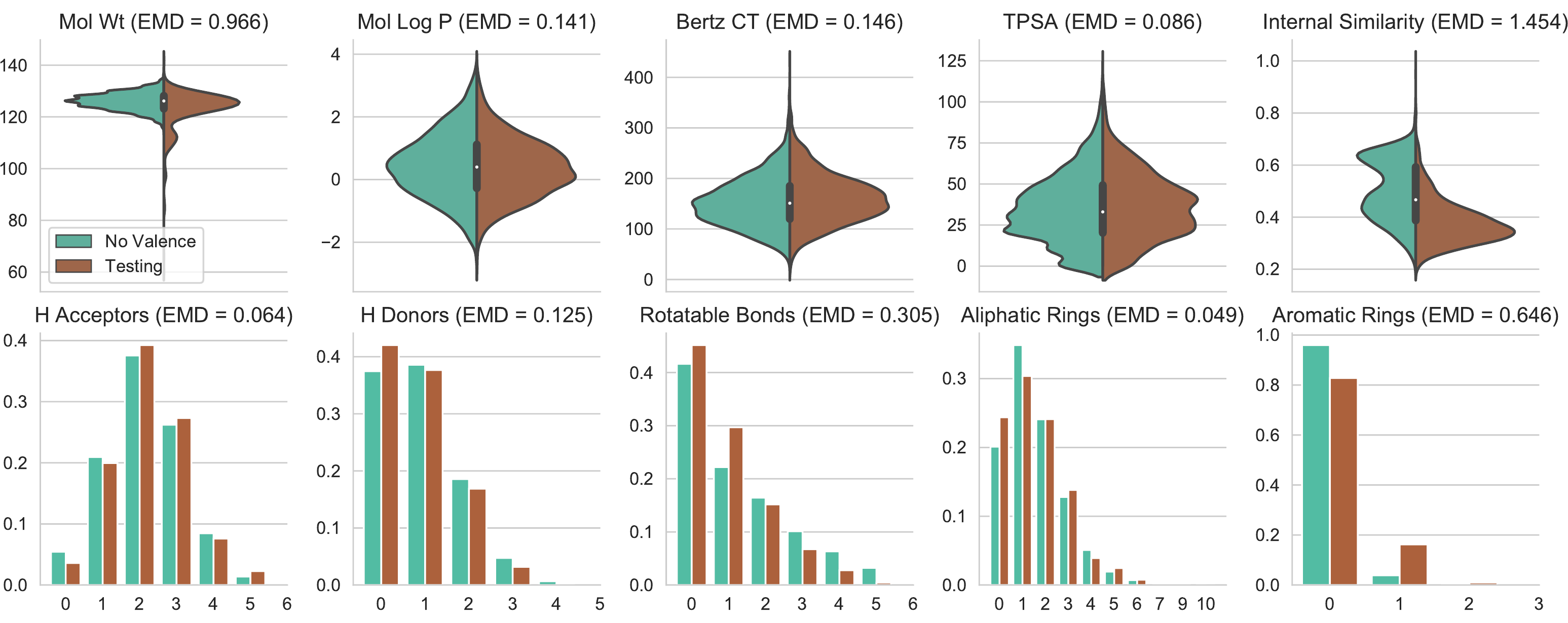}%
    \end{minipage}\vfill%
    \begin{minipage}[t][][b]{0.03\textwidth}{\scriptsize(c)}\end{minipage}%
    \begin{minipage}[t][][b]{0.85\textwidth}%
        \includegraphics[width=\linewidth]{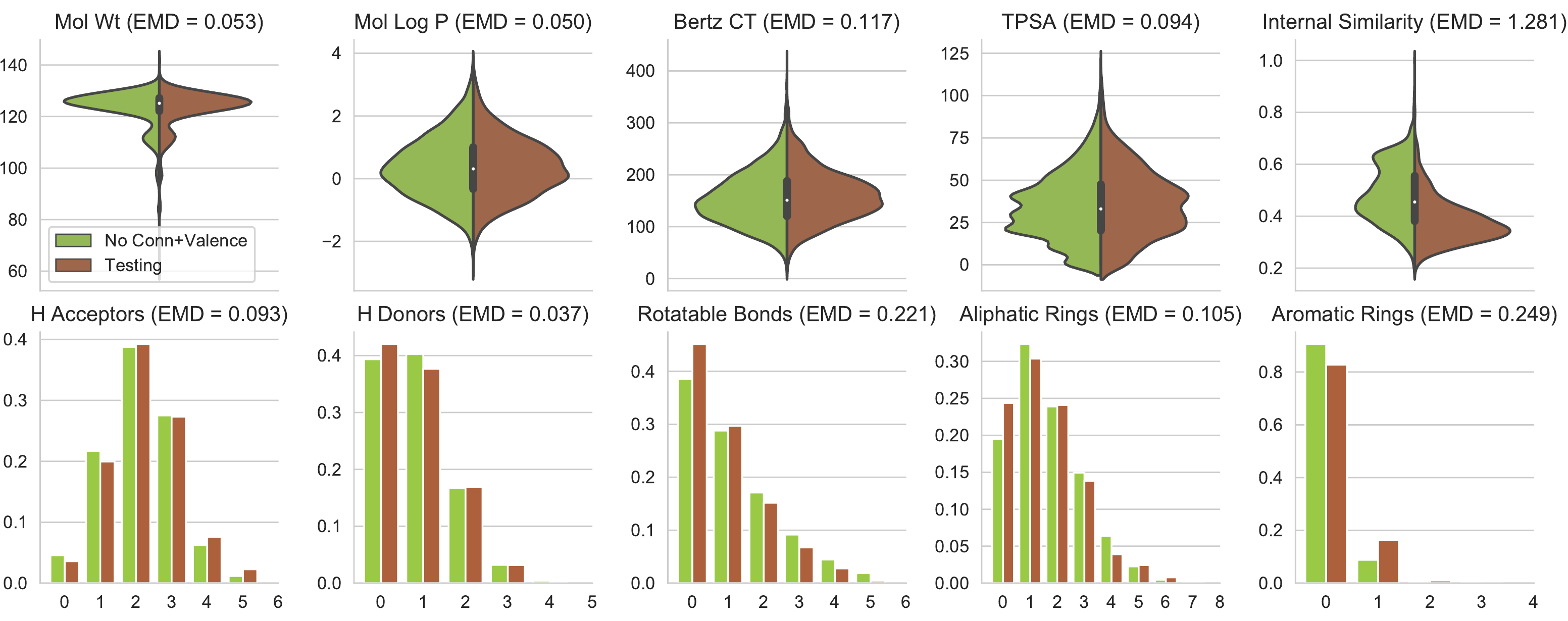}%
    \end{minipage}%
\end{figure}
\begin{figure}[h]
    \centering
    \begin{minipage}[t][][b]{0.03\textwidth}{\scriptsize(d)}\end{minipage}%
    \begin{minipage}[t][][b]{0.85\textwidth}%
        \includegraphics[width=\linewidth]{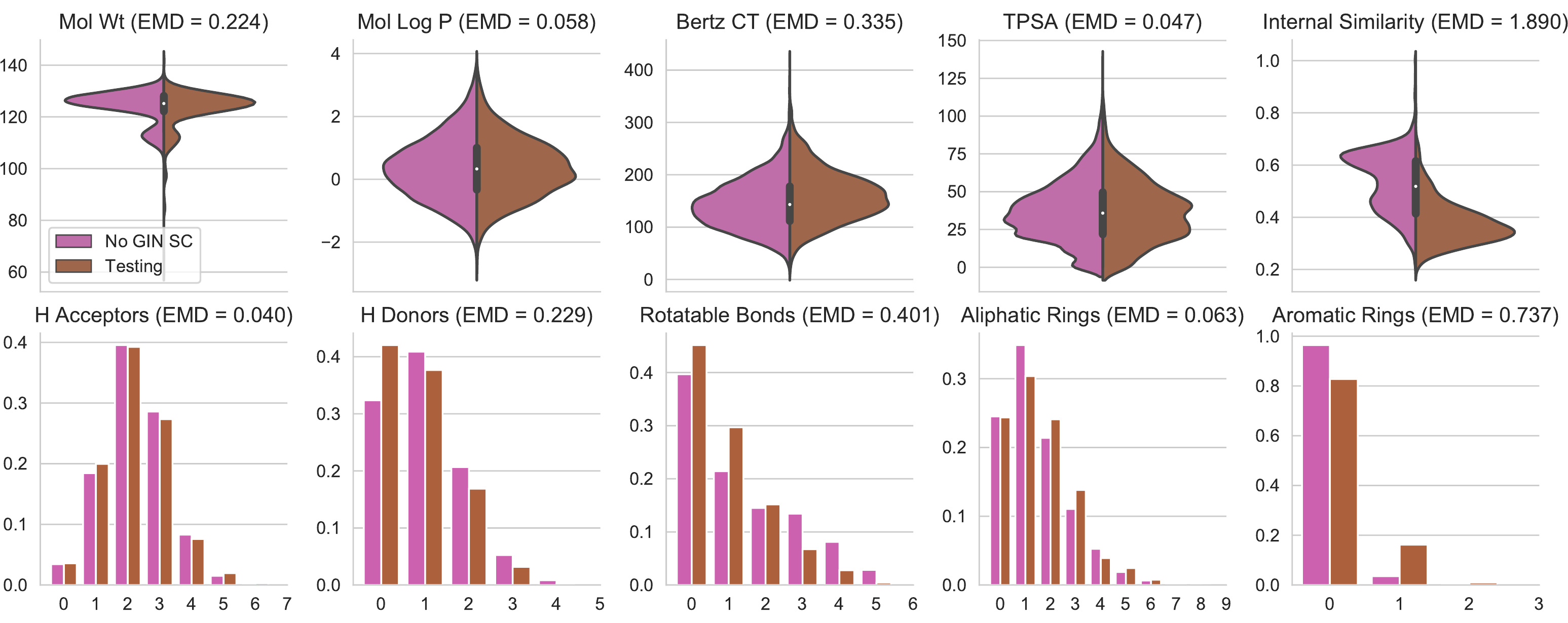}%
    \end{minipage}\vfill%
    \begin{minipage}[t][][b]{0.03\textwidth}{\scriptsize(e)}\end{minipage}%
    \begin{minipage}[t][][b]{0.85\textwidth}%
        \includegraphics[width=\linewidth]{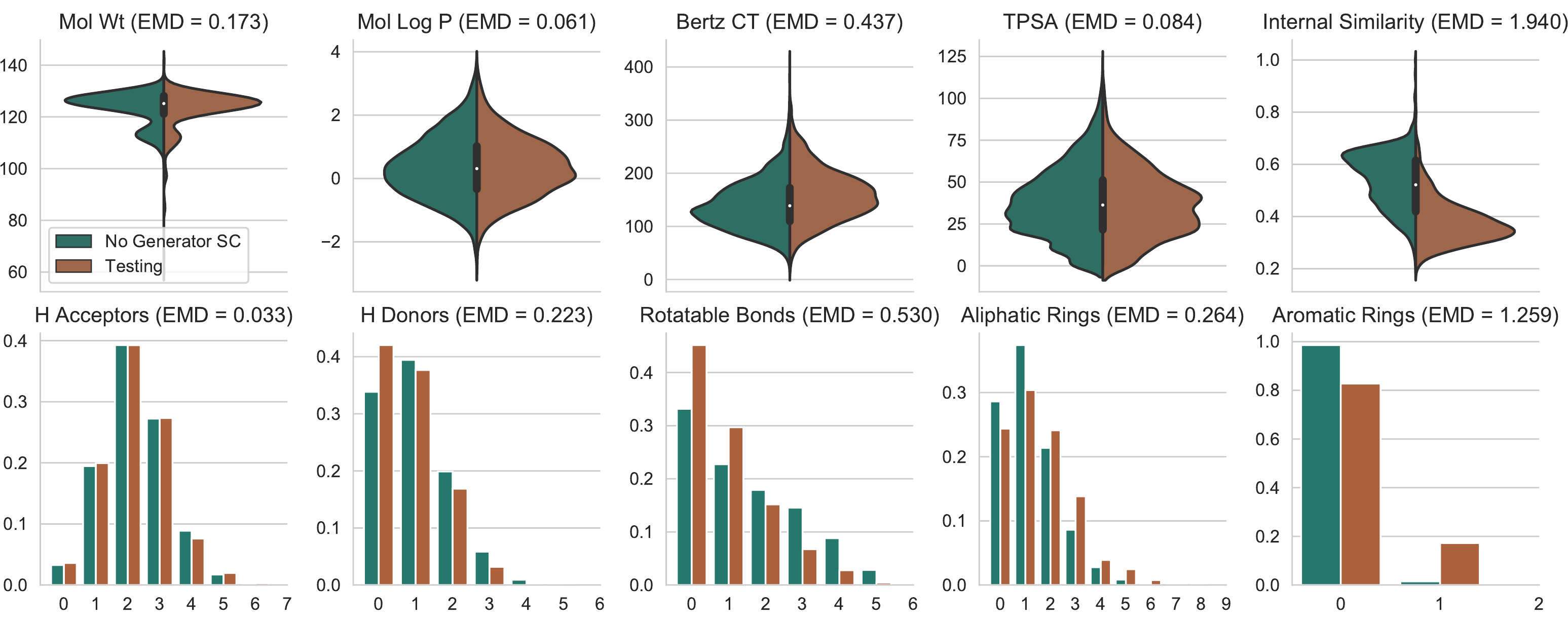}%
    \end{minipage}\vfill%
    \begin{minipage}[t][][b]{0.03\textwidth}{\scriptsize(f)}\end{minipage}%
    \begin{minipage}[t][][b]{0.85\textwidth}%
        \includegraphics[width=\linewidth]{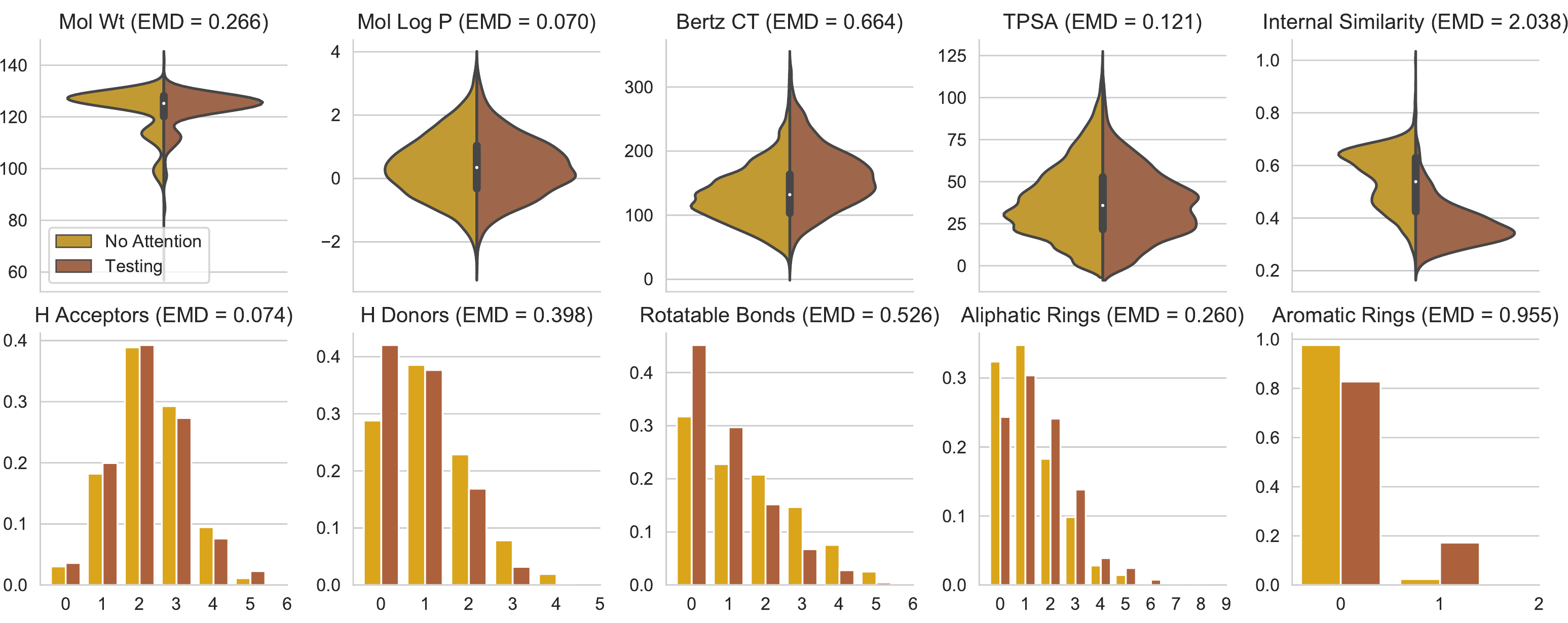}%
    \end{minipage}\vfill%
    \begin{minipage}[t][][b]{0.03\textwidth}{\scriptsize(g)}\end{minipage}%
    \begin{minipage}[t][][b]{0.85\textwidth}%
        \includegraphics[width=\linewidth]{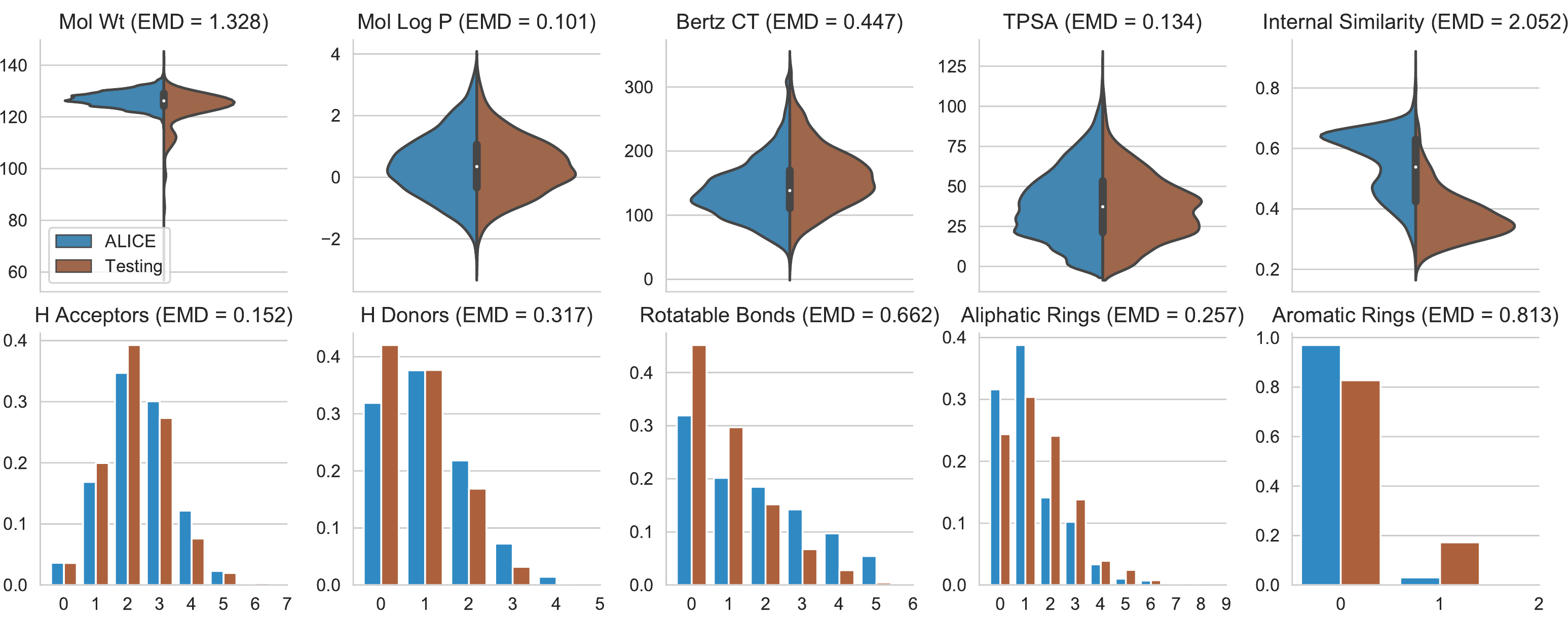}%
    \end{minipage}%
\end{figure}
\begin{figure}[t]
    \centering
    \begin{minipage}[t][][b]{0.03\textwidth}{\scriptsize(h)}\end{minipage}%
    \begin{minipage}[t][][b]{0.85\textwidth}%
        \includegraphics[width=\linewidth]{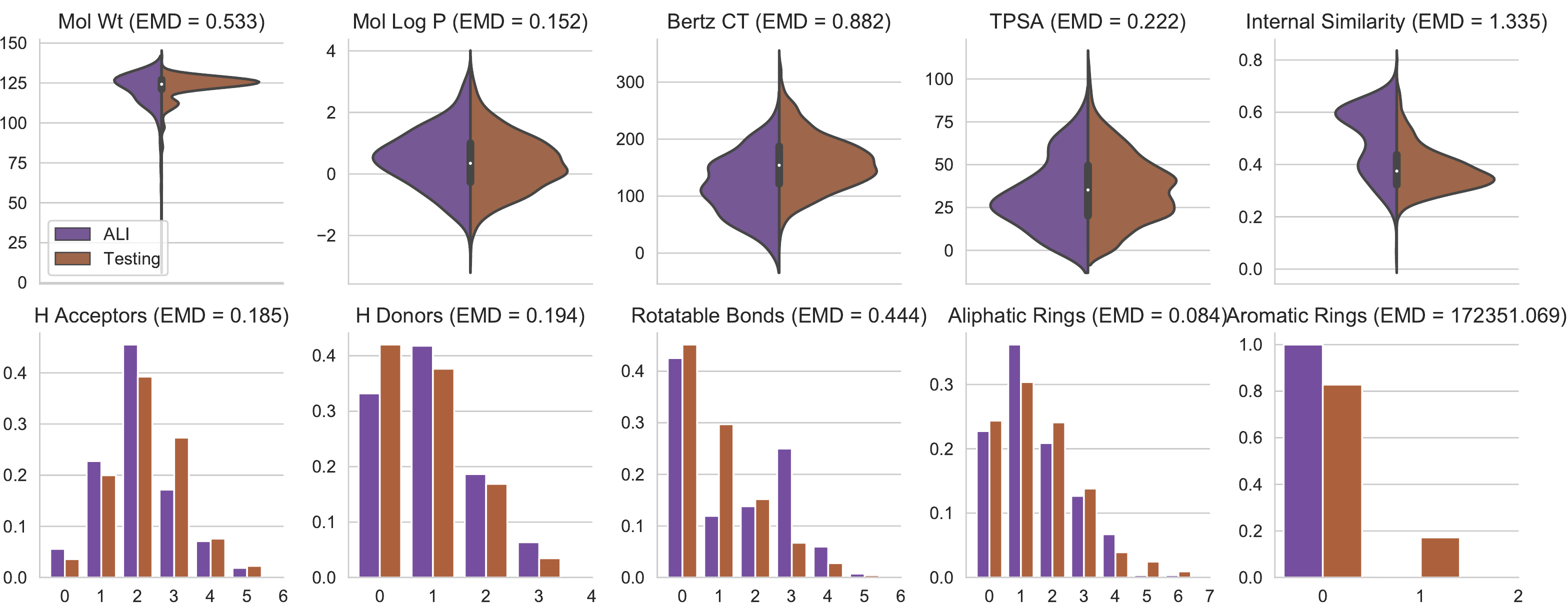}%
    \end{minipage}\vfill%
    \begin{minipage}[t][][b]{0.03\textwidth}{\scriptsize(i)}\end{minipage}%
    \begin{minipage}[t][][b]{0.85\textwidth}%
        \includegraphics[width=\linewidth]{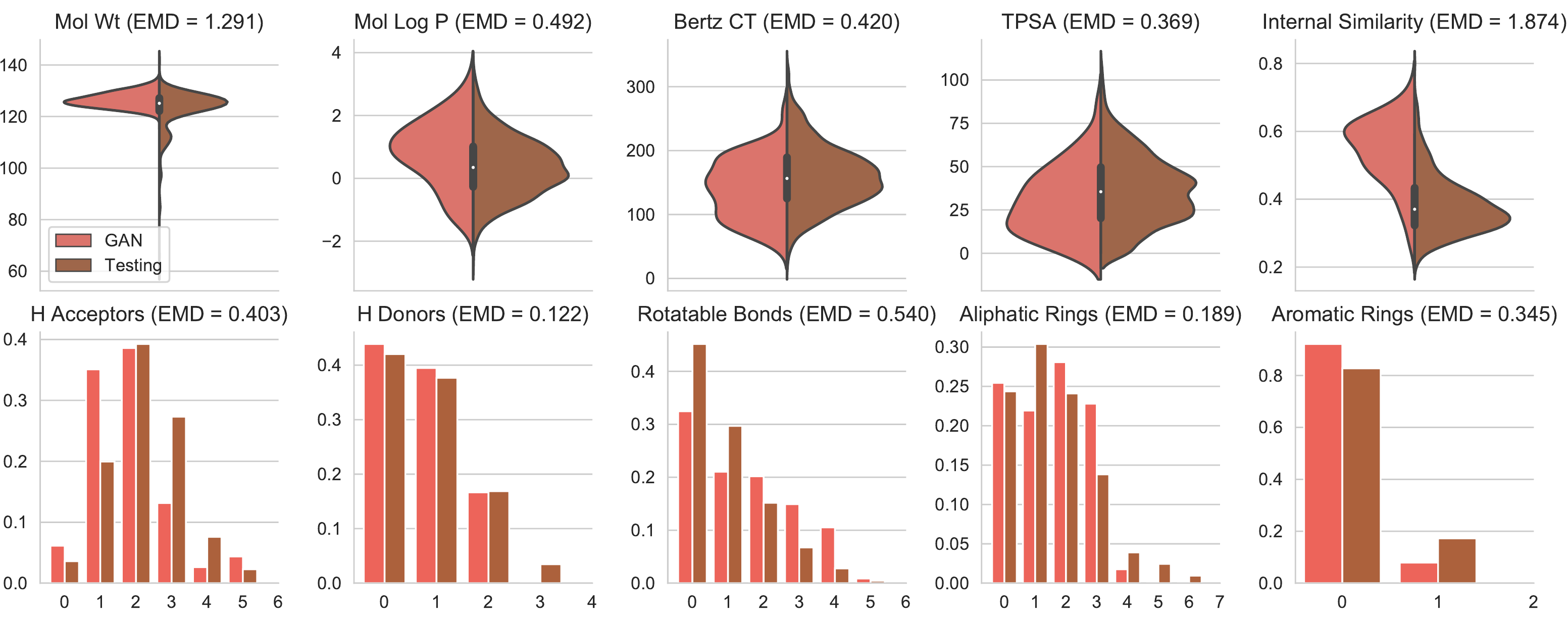}%
    \end{minipage}\vfill%
    \begin{minipage}[t][][b]{0.03\textwidth}{\scriptsize(j)}\end{minipage}%
    \begin{minipage}[t][][b]{0.85\textwidth}%
        \includegraphics[width=\linewidth]{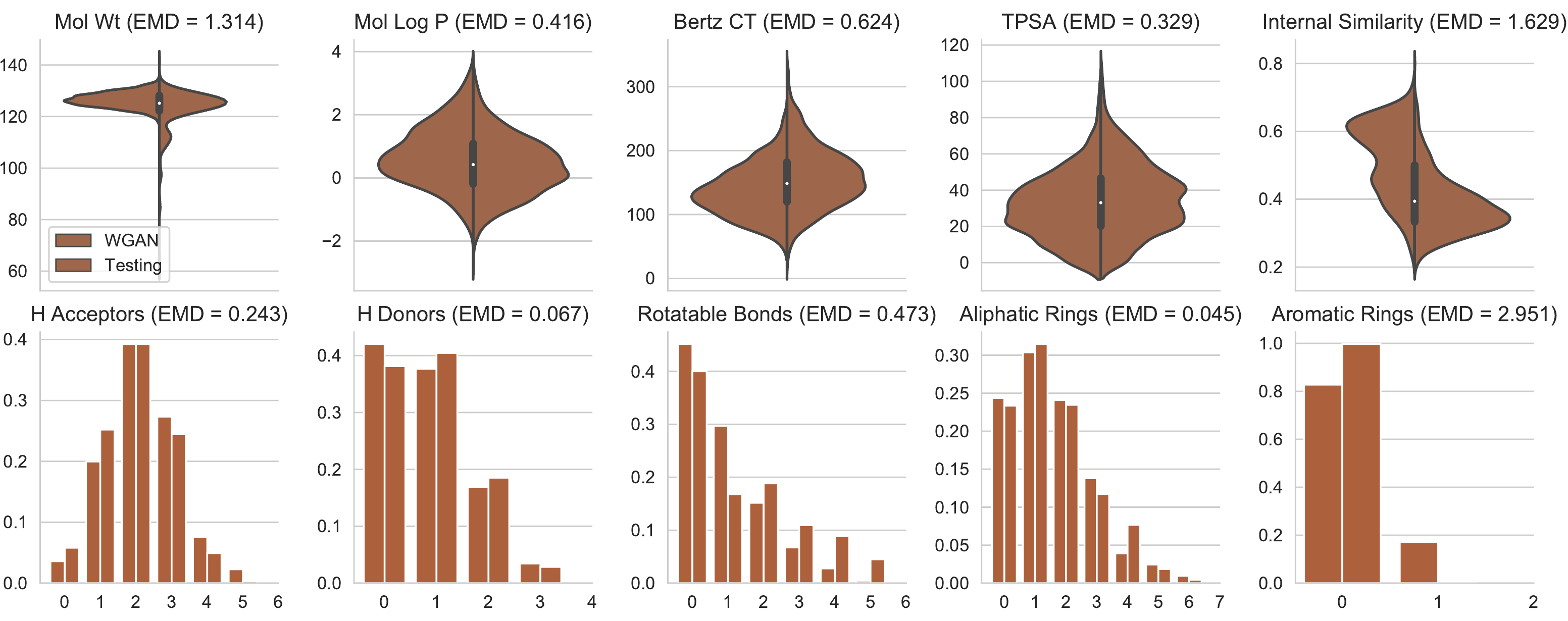}%
    \end{minipage}
\end{figure}
\clearpage

\section{Implementation Details}\label{sec:supp_implementation}

\begin{table}[tb]
    \centering
    \caption{Number of learnable weights for each model.}
    \begin{tabular}{lr}
        \toprule
        Model & Parameters \\
        \midrule
        \ourmodel\ & 1.11M \\  %
        \quad No Encoder Skip-Conn. & 1.03M \\  %
        \quad No Generator Skip-Conn. & 1.10M \\  %
        \quad No Attention & 1.03M \\  %
        \quad ALICE & 987K \\  %
        \quad ALI & 729K \\  %
        \quad (W)GAN & 509K \\  %
        CGVAE & 13.6M \\
        MolGAN & 439K \\
        NeVAE & 0.5K \\
        GrammarVAE & 5.4M \\
        \bottomrule
    \end{tabular}
\end{table}

\paragraph{GIN.}
We implemented the sum over the descriptors
of neighboring nodes in the GIN layer \eqref{eq:gin-layer}
as a dense matrix product between the adjacency
matrix $\mathbf{A}_{\bullet,\bullet,r_k}$
and the matrix
$(\mathbf{h}_{v_1}^{(l)}, \ldots, \mathbf{h}_{v_n}^{(l)})^\top$,
which has a complexity of $O(n^2 d)$.
Clearly, this is a bottleneck for large graphs.
However, due to the sparsity of the adjacency matrix,
we can be employ sparse matrix computation.
In the first layer, where descriptors are one-hot encoded
node types, the computation requires $O(\bar{e})$ operations,
with $\bar{e}$ being the average number of edges between nodes.
After the second layer, descriptors will be dense and
we have to compute the product between a sparse and a dense
matrix, which requires $O(\bar{e}d)$ operations.
To the best of our knowledge, most deep learning
frameworks only support sparse computation on rank-2 matrices,
therefore additional engineering would be required to
incorporate the batch dimension for efficient sparse matrix
computation.

\paragraph{Architecture.}
If not mentioned otherwise, tanh is used as activation function.
Encoders and discriminators have $L=2$ GIN-layers with 128 units
for each of the $m=3$ edge types.
The GIN-MLP in \eqref{eq:gin-layer} has one linear layer.
The two linear layers used for soft-attention and
subsequent graph-pooling in \eqref{eq:soft-attention-pooling}
have 128 units each.
The MLP taking the graph-pooled descriptor $\mathbf{h}_G$
and outputting $\mathbf{h}_G^\prime$ has two fully-connected
layers with 128 and 64 units, respectively.
The cycle-discriminator $D_\eta(G_1, G_2)$ uses a 2-layer MLP
with 64 units each to combine graph-level feature descriptors
$\mathbf{h}_{G_1}^\prime$ and $\mathbf{h}_{G_2}^\prime$.
Networks $D_\psi(G, \mathbf{z})$, $g_\phi(G, \varepsilon)$ have
a 96-dimensional noise vector as second input, which is the input to one
fully-connected layer with 256 units.
Likewise, $g_\theta(\mathbf{z}, \varepsilon)$ takes two 96-dimensional vectors;
$\mathbf{z}$ is split into three 32-dimensional vectors used in the decoder's skip
connections and $\varepsilon$ is concatenated with $\mathbf{z}_1$ and
fed to a 3-layer MLP with 128, 256, and 512 units.
The hyper-parameter $\tau$ used in the Gumbel-softmax \cite{Jang2017,Maddison2017}
is set to $\tau = 1$.
We trained for 250 epochs using the Adam optimizer \cite{Kingma2014} with $\beta_1 = 0.5$, $\beta_2 = 0.9$,
with initial learning rates 0.001 and 0.0004
for the generator and discriminator, respectively.
We lowered the initial learning rates by a factor of 0.5, 0.1, and 0.01
after 80, 150, and 200 epochs, respectively.
The regularization weights $\mu$ and $\nu$ in \eqref{eq:connectivity-penalty} and
\eqref{eq:valence-penalty} were tuned manually
and set to $0.005$ and $0.05$.
In practice, we compute a smoothed version of $\tilde{\mathbf{B}}_{ij}$
in \eqref{eq:num-paths} using $s(x) = \sigma(a(x - \frac{1}{2}))$
with $a = 100$, which improves numerical stability.
The weight of the gradient penalty \cite{Gulrajani2017} of the
discriminators was set to 10.
Weights $\theta$ and $\phi$ of the encoder and decoder
get updated at the same time, as are the weights $\psi$, $\eta$, and $\xi$
of the discriminators.
Updates are performed sequentially, with a
1:1 ratio of discriminator to encoder/decoder updates.
To stabilize training and encourage exploration, we added a discount
term proportional to the variance of the predictions of the
discriminator $D_\psi(\tilde{G}, \mathbf{z})$. The term was weighted by $-0.2$
and added to the remaining losses when updating
weights $\theta$ and $\phi$ of the encoder and decoder.

\paragraph{Random Graph Generation.}

For the random graph generation model, we used a similar setup as the
generator used in NeVAE~\cite{Samanta2018}.
We first randomly select the atom types for each node in the graph
to create $\mathbf{X}$, and then sequentially draw edges $(v_i, r_k, v_j)$ such
that the valence of atoms is always correct.
We denote by $\delta(i, j\,|\,\mathcal{E}_{l}) \in \{0; 1\}$ whether
nodes $v_i$ and $v_j$ are disconnected in the graph
specified by the set of edges $\mathcal{E}_{l}$,
and by $\delta(r_k\,|\,v_i, v_j, \mathcal{E}_{l}) \in \{0; 1\}$
whether nodes $v_i$ and $v_j$ can be connected by an
edge of type $r_k$ without violating valence constraints.
The first mask is used to ensure only a single bond between atoms is formed,
and the second mask to satisfy valence constraints.
Independent random noise drawn from $\mathcal{N}(0,1)$ is denoted by $\varepsilon$.

The probability of the $i$-th node feature vector $\mathbf{x}_{v_i}$
encoding the atom type is given by
$$
P(\mathbf{x}_{v_i} = \mathbf{e}_j)
= \frac{\exp{\varepsilon_j}}{\sum_{j^\prime=1}^d \exp{\varepsilon_{j^\prime}}} ,
$$
where $\mathbf{e}_j$ is the one-hot encoding of the $j$-th atom type
($j=1,\ldots,d$).
Edges are added sequentially: in the $l$-th step,
the probability of an edge $(v_i, r_k, v_j)$,
conditional on the previously added edges $\mathcal{E}_{l-1}$
is given by
\begin{align*}
P((v_i, r_k, v_j)\,|\, \mathcal{E}_{l-1})
&= P(v_i, v_j\,|\, \mathcal{E}_{l-1})
P(r_k\,|\, v_i, v_j, \mathcal{E}_{l-1}), \\
P(v_i, v_j\,|\, \mathcal{E}_{l-1}) &=
\frac{ \delta(i, j\,|\,\mathcal{E}_{l-1}) \exp(\varepsilon_{i,j}) }
{ \sum_{i^\prime, j^\prime} \delta(i^\prime, j^\prime\,|\,\mathcal{E}_{l-1}) \exp(\varepsilon_{i^\prime, j^\prime}) }, \\
P(r_k\,|\, v_i, v_j, \mathcal{E}_{l-1}) &=
\frac{ \delta(r_k\,|\,v_i, v_j, \mathcal{E}_{l-1}) \exp(\varepsilon_{r_k}) }
{ \sum_{k^\prime=1}^r \delta(r_{k^\prime}\,|\,v_i, v_j, \mathcal{E}_{l-1}) \exp(\varepsilon_{r_{k^\prime}}) } .
\end{align*}
We repeatedly sample edges until no additional valid edges can be added.
Therefore, generated random molecular graphs will always have correct valences, but
can have multiple connected components
if valence constraints cannot be satisfied otherwise; we consider
these samples to be invalid.

\paragraph{Software.}
We implemented our model using TensorFlow version 1.10.0
and performed training on a NVIDIA Quadro P6000.
We used RDKit \cite{Landrum2018} version 2018.09.3.0
for computing molecular descriptors.
Validity of generated molecular graphs was assessed
by RDKit's \texttt{SanitizeMol} function.
To evaluate how well models estimate the distribution of molecules
from the training data, we used the benchmark suite
GuacaMol version 0.3.2 \cite{Brown2018}
and the Python Optimal Transport (POT) library version 0.5.1 \cite{flamary2017pot}.

\end{document}